\documentclass[10pt,twocolumn,letterpaper]{article}

\usepackage{cvpr}%
\usepackage{graphicx}
\usepackage{amsmath}
\usepackage{amssymb}
\usepackage{booktabs}
\usepackage{graphicx}
\usepackage{caption}
\usepackage{multirow}
\usepackage{float}
\usepackage{array}
\usepackage[accsupp]{axessibility}  %
\newcolumntype{P}[1]{>{\centering\arraybackslash}p{#1}}
\newcolumntype{M}[1]{>{\centering\arraybackslash}m{#1}}

\usepackage[pagebackref,breaklinks,colorlinks]{hyperref}

\usepackage[capitalize]{cleveref}
\crefname{section}{Sec.}{Secs.}
\Crefname{section}{Section}{Sections}
\Crefname{table}{Table}{Tables}
\crefname{table}{Tab.}{Tabs.}

\def\cvprPaperID{27} %
\def\confName{CVPR}
\def\confYear{2023}

\newcommand*{\affaddr}[1]{#1} %
\newcommand*{\affmark}[1][*]{\textsuperscript{#1}}
\newcommand*{\email}[1]{\texttt{#1}}
\begin{document}

\title{Density Invariant Contrast Maximization for Neuromorphic Earth Observations}

\author{%
Sami Arja\affmark[*,1], Alexandre Marcireau \affmark[1], Richard L. Balthazor\affmark[2] \\ Matthew G. McHarg\affmark[2], Saeed Afshar\affmark[1] and Gregory Cohen\affmark[1]\\
\rule{0pt}{3ex} \affaddr{\affmark[1]Western Sydney University}\\
\affaddr{\affmark[2]United States Air Force Academy}\\
\email{$^{*}$s.elarja@westernsydney.edu.au}\\
}

\maketitle

\begin{abstract}   
   Contrast maximization (CMax) techniques are widely used in event-based vision systems to estimate the motion parameters of the camera and generate high-contrast images. However, these techniques are noise-intolerance and suffer from the multiple extrema problem which arises when the scene contains more noisy events than structure, causing the contrast to be higher at multiple locations. This makes the task of estimating the camera motion extremely challenging, which is a problem for neuromorphic earth observation, because, without a proper estimation of the motion parameters, it is not possible to generate a map with high contrast, causing important details to be lost. Similar methods that use CMax addressed this problem by changing or augmenting the objective function to enable it to converge to the correct motion parameters. Our proposed solution overcomes the multiple extrema and noise-intolerance problems by correcting the warped event before calculating the contrast and offers the following advantages: it does not depend on the event data, it does not require a prior about the camera motion, and keeps the rest of the CMax pipeline unchanged. This is to ensure that the contrast is only high around the correct motion parameters. Our approach enables the creation of better motion-compensated maps through an analytical compensation technique using a novel dataset from the International Space Station (ISS). Code is available at \url{https://github.com/neuromorphicsystems/event_warping}
\end{abstract}

\section{Introduction}
\label{sec:intro}

\begin{figure}[ht]
  \centering
  \includegraphics[width=3in]{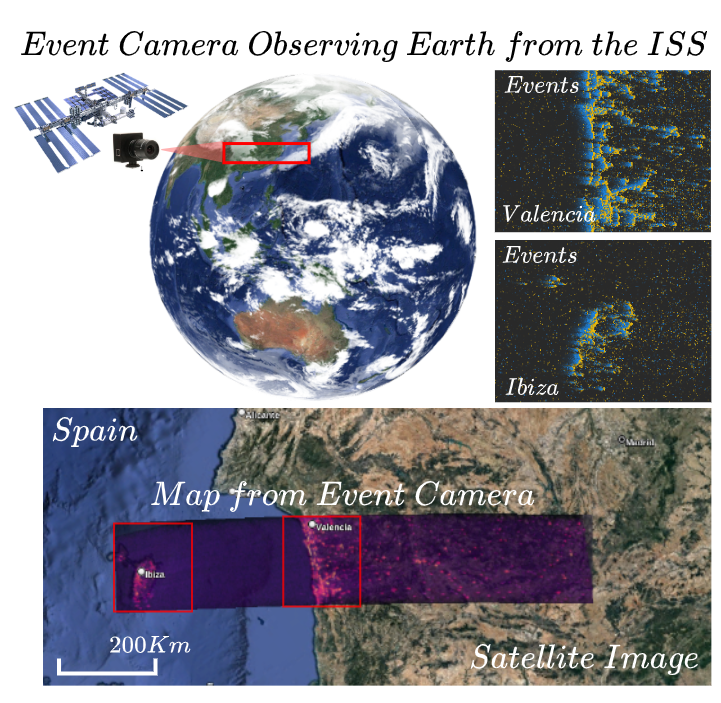}
  \caption{Neuromorphic Earth Observations application. The Falcon Neuro project  placed two event cameras on the ISS in 2021. These sensors have been used for earth observation and have captured a dataset of a variety of different locations in 2022. To produce panoramic map images from the event-based output of these sensors, a contrast maximization algorithm, such as CMax \cite{gallego_unifying_2018}, is needed to compensate for the motion of the ISS. In this paper, we proposed an analytical approach that creates high-contrast panoramic images using CMax \cite{gallego_unifying_2018} by making the algorithm invariant to the density of events by leveraging the physical properties of the events and their geometries. This was performed by introducing a geometric piecewise correction function that adjusts the warped image to prevent the loss landscape from forming multiple extrema.}
  \label{fig:problemsolution}
\end{figure}

Neuromorphic event cameras \cite{lichtsteiner_128times128_2008,Finateu2020510A1} are biology-inspired optical sensors that offer high-speed, high-dynamic range, and low data rate operation, making them extremely well suited for use in the space environment. The sensors are asynchronous and have in-pixel circuitry to produce high-temporal contrast change events only in response to changes in the visual scene. The change events can be represented as event $e=(\boldsymbol{u},p,t)$, where $\boldsymbol{u}=(x,y)$ is the pixel coordinates of the event occurrence, $p$ is the polarity of the contrast change, corresponding to whether the change in the brightness in increasing or decreasing, $t$ is the timestamp of the change in the order of $\mu s$.

The asynchronous nature of the events provides numerous advantages over traditional vision sensors, such as superior dynamic range, low latency, high temporal resolution, and significantly lower power consumption. The high temporal resolution removes the effects of motion blur, and the change detection suppresses redundant information at the pixel level. These features make the sensors well suited to tackling challenging machine vision tasks such as recognition \cite{bethi_optimised_2022,afshar_event-based_2020,sironi_hats_2018,lagorce_hots_2017,orchard_hfirst_2015}, tracking \cite{ralph_real-time_2022,ni_visual_2015,gehrig_eklt_2020,mueggler_event-based_2014,tedaldi_feature_2016,glover_robust_2017,alzugaray_asynchronous_2018,kueng_low-latency_2016,zhu_event-based_2017}, SLAM \cite{Kim2016RealTime3R,Gehrig_2020_CVPR,Rebecq18ijcv,hutchison_simultaneous_2013,kim_simultaneous_2014,vidal_ultimate_2018}, motion estimation \cite{gallego_unifying_2018,willert_event-based_2022,gallego_accurate_2017,ozawa_accuracy_2022,gallego_focus_2019,stoffregen_event_2019,ozawa_recursive_2022,shiba_fast_2023,nunes_robust_2022,seok_robust_2020,liu_globally_2020,shiba_event_2022,mcleod_globally_2022,peng_globally-optimal_2021,liu_spatiotemporal_2021}, and space domain awareness and space imaging \cite{cohen2018approaches,afshar_event-based_2020,cohen_event-based_2019,ralph_astrometric_2022,kaminski_observational_2019}. The lack of a frame-based output  requires the development of new algorithms and systems, and can allow for modes of operation and sensing not possible with conventional imaging sensors.

The recent advances in event-based algorithms and the wide availability of high-resolution event cameras \cite{Finateu2020510A1} have led to wide adoption in numerous real-world research applications such as in Autonomous Underwater Vehicles (AUV) \cite{zhang_event-based_2022}, ground-based mobile telescopes \cite{cohen2018approaches,cohen_event-based_2019}, Unmanned Aerial Vehicles (UAV) \cite{vitale2021event,ramesh2018long,mueggler2017event}, ground robots and vehicles \cite{milde_obstacle_2017,iaboni_event_2021,maqueda_event-based_2018}, and even in space onboard the ISS \cite{mcharg_falcon_2022}.

There is increasing interest in using event cameras in the space environment and further afield. These include the investigation of their use in future lunar spacecraft landing tasks \cite{sikorski_event-based_2021,mcleod_globally_2022} and for underground exploration using the Mars Ingenuity Helicopter \cite{mahlknecht_exploring_2022}. These projects used either simulated events from video sequences or earth-based environments that resemble the Martian and the Lunar surfaces. 

In this paper, we investigate neuromorphic earth observation using event cameras mounted on the ISS. These sensors were installed in 2021 as part of a collaboration between the United States Air Force Academy and Western Sydney University through the Falcon Neuro project \cite{mcharg_falcon_2022}. The Falcon Neuro payload contains two identical neuromorphic sensors, with one pointed in the RAM direction, and the other pointed in the NADIR direction. This work focuses on techniques to process data from the NADIR camera to produce visual maps of the earth through an analytical solution to the original CMax framework.

\subsection{Motivation}
\label{sec:motiv}

A state-of-the-art approach for motion estimation was first introduced by \cite{gallego_unifying_2018}, known as CMax. It works by estimating the camera's relative motion vector, $\boldsymbol{\theta}=[v_x,v_y]$, over a time window $\delta = t_i-t_{ref}$ to align events with the edges or objects that generated them. This involves adjusting the pixel coordinates of individual events to eliminate the motion induced by the sensor or object to create sharp images. Specifically, each event $e_i=(\boldsymbol{u_i},p_i,t_i)$ is warped by a shear transformation based on a motion candidate $\boldsymbol{\theta}$

\begin{equation}
    \boldsymbol{u'_i}=\left[ \begin{matrix}
    x'_i \\ y'_i
    \end{matrix} \right]=\left[ \begin{matrix}
    x_i \\ y_i
    \end{matrix} \right]-\left[ \begin{matrix}
    v_x \\ v_y
    \end{matrix} \right]*\delta \label{eq:warpingopt}
\end{equation}

This works by reversing the motion $\boldsymbol{\theta}$ between $t_i$ and the beginning of $\delta$ and changes the spatial location of the $\boldsymbol{u_i}$. The new events are then accumulated into an image $H$ or also called Image of Warped Events (IWE) as in (\ref{eq:accumulate}). Each pixel in (\ref{eq:accumulate}) sums the values of the warped events $u^{'}_i$ that fall within it.

\begin{equation}
    H(\boldsymbol{u'};\boldsymbol{\theta}) \dot{=}\sum_{i=1}^{N}b_{k}\delta(\boldsymbol{u}-\boldsymbol{u'_{i}}),\label{eq:accumulate}
\end{equation}

\hspace{-0.5cm} where $b_{k}$ is the number of events along the trajectories as detailed in \cite{gallego_unifying_2018}. The contrast of $H$ is then calculated as a function of $\boldsymbol{\theta}$

\begin{equation}
    C(\boldsymbol{\theta})\dot{=}\frac{1}{N}\sum_{j=1}^{N}( H(\boldsymbol{u'_j};\boldsymbol{\theta})-\mu(\boldsymbol{\theta}))^{2}, \label{eq:variance}
\end{equation}

\hspace{-0.5cm} where N is the number of pixels in $H$, and $\mu(\boldsymbol{\theta})$ is the mean of the pixel intensity of $H$. The strategy is to find the correct $\boldsymbol{\theta}$ that maximises the objective function $C(\boldsymbol{\theta})$, in this case, a higher $C(\boldsymbol{\theta})$ indicates higher events alignment and a sharp motion-corrected image $H$.

Despite the recent successes of the CMax algorithm in various applications, the algorithm suffers from a few fundamental weaknesses particularly, in space-based earth mapping applications caused by the increased density of events resulting from the continuous movement of the camera in the orbit. Below we discuss these limitations and use them to motivate the method proposed in this paper. 

\begin{figure}[t]
  \hspace*{-0.5cm}
  \centering
  \includegraphics[width=3.5in]{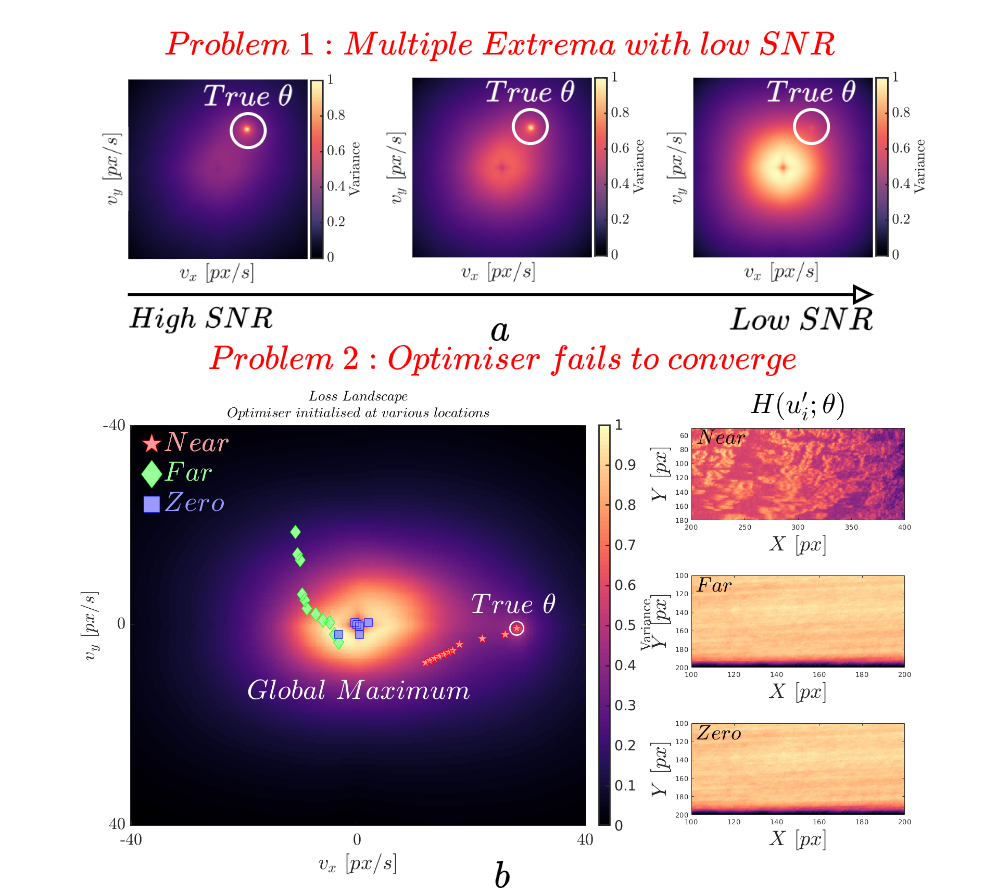}
  \caption{Illustration of CMax problems in low SNR scenes. \textbf{Problem 1:} It is trivial to estimate the true motion $\boldsymbol{\theta}$ in high SNR scenes, however, as the SNR decreases the contrast around the wrong $\boldsymbol{\theta}$ increases significantly, overtaking the true value. \textbf{Problem 2:} Optimisation algorithm converges only when it is initialised near the true $\boldsymbol{\theta}$ and fails everywhere else, this is a problem because a robust CMax should converge to the correct motion $\boldsymbol{\theta}$ without any prior.}
   \label{fig:motivationproblems}
\end{figure}

This work makes use of the event camera pointed directly downward (the NADIR direction) as shown in Figure \ref{fig:problemsolution}. The ISS moves with a consistent speed of 17,900 mph (i.e. 8 km/s) \cite{issfacts} and the movement can be modeled as consisting primarily of translation, except during docking procedures. The number of events recorded by the camera is predominantly influenced by the texture of the surface of the earth and the variations in lighting conditions over the day or night side of the earth.

The high texture of the lower atmosphere (e.g. clouds) combined with pixel noise caused by the camera setting and circuit mismatch, reduces the Signal-to-Noise Ratio (SNR). A low SNR scene results in fewer structures captured by the camera, affecting the accuracy of the motion estimation $\boldsymbol{\theta}$. When the density of events in (\ref{eq:accumulate}) increases, this also leads to higher variance in (\ref{eq:variance}) around the wrong $\boldsymbol{\theta}$. As a consequence, higher contrast does not always imply better events alignment, as illustrated in Figure \ref{fig:motivationproblems}(a), where multiple extrema become visible. This noise intolerance issue was initially identified by \cite{stoffregen_event_2019} but was not further investigated.

Recent studies have shown that maximization (\ref{eq:variance}) can be done either by using a conjugate gradient \cite{gallego_unifying_2018,gallego_accurate_2017} approach or a branch-and-bound \cite{liu_globally_2020} method. The former requires a good initialisation of $\boldsymbol{\theta}$ to converge to the correct local minima, as summarised in Figure \ref{fig:motivationproblems}(b), and the latter is a global optimisation and search method better suited for rotation motion estimation. In addition, recent improvements have included refining the objective function to better suit the targeted settings \cite{stoffregen_event_2019,gallego_focus_2019}. However, this makes the objective function and optimisation application-dependent and may increase complexity.

Guided by these observations, we propose a new approach that corrects (\ref{eq:accumulate}) by only considering the motion and geometry of the warped events, which modifies the landscape of the objective function in (\ref{eq:variance}) automatically. This modification enables us to keep the optimisation algorithm and objective function unchanged in a way invariant to the nature of the input data.

\subsection{Contribution}
\label{sec:contrib}
We present a novel approach that enables space-based earth mapping using the CMax algorithm. Our approach makes the following assumptions: the speed of the camera is constant, the time window $\delta$ is the entire event stream and no motion prior is given to the algorithm. Our method not only provides an optimal solution to this specific application, but it also addresses several fundamental problems, including determining the appropriate objective function to use \cite{gallego_focus_2019} and deciding how many events to process at once \cite{stoffregen_event_2019}. We demonstrate that the variance can serve as an optimal objective function without any modifications, and all events can be utilised in the case of translation motion. While our method does not address other types of motion, such as rotation and zooming, this eliminates the need to test multiple objective functions and employ several batch sizes to estimate the optimal motion parameters.

Our method relies solely on the overall geometry of the warped events and it does not depend on noise density or the size of the time window. It enables CMax to consistently produce high-contrast outputs to equation (\ref{eq:variance}) around the correct $\boldsymbol{\theta}$, even in cases where noise dominates the primary structure of the scene. This is accomplished by increasing the value of specific pixels of the Image of Warped Events – namely, pixels that correspond to parts of the scene in front of which the sensor spent less time. Our approach recovers a single solution for $\boldsymbol{\theta}$. This significantly increases the rate of convergence to the correct solution using a simple optimisation search method such Nelder-Mead, thus overcoming the problem of multiple extrema.

We model the noise events of noise-rich scenes as a uniform rectangular cuboid in the 3-dimensional space $\left\{x, y, t\right\}$. This allows us to analytically calculate the variance of the accumulated image, by shearing the cuboid and integrating the function that describes its height after shearing.

Our method was evaluated on a diverse dataset captured from the ISS. The results were assessed both qualitatively and quantitatively using Root Mean Square (RMS) error and the Rate of Convergence (RoC) metrics for evaluation - see Section \ref{sec:quality} and \ref{sec:quatity}.

\subsection{Related Work}\label{sec:relatedwork}

The CMax framework was introduced by \cite{gallego_unifying_2018,gallego_accurate_2017} and its performance was further investigated in \cite{gallego_focus_2019,stoffregen_event_2019}. Different alignment methods have been proposed to leverage the benefit of this algorithm \cite{nunes_robust_2022,Gu21iccv,seok_robust_2020} for various tasks such as rotational motion estimation \cite{gallego_accurate_2017,kim_simultaneous_2014,reinbacher_real-time_2017,kim_real-time_2021}, optical flow \cite{zhu2019unsupervised,hagenaarsparedesvalles2021ssl,Shiba22eccv}, 3D reconstruction \cite{rebecq_emvs_2016,ghosh_multieventcamera_2022}, depth estimation \cite{gallego_unifying_2018}, motion segmentation \cite{Stoffregen19iccv,Zhou21tnnls,parameshwara_0-mms_2021}, and intensity reconstruction \cite{Zhang22pami}. To estimate the motion, existing works either rely on local \cite{gallego_unifying_2018,gallego_focus_2019,Stoffregen19iccv,stoffregen_event_2019}, or global optimisation \cite{liu_globally_2020,peng_globally-optimal_2021} to facilitate the convergence to the correct motion parameters.

\cite{shiba_fast_2023} is the work most closely related to ours. Their method augments the objective function and applies corrections derived from mathematical models. However, \cite{shiba_fast_2023} tackles the problem of event collapse, which occurs when objects are not moving parallel to the sensor plane. By contrast, we only focus on translational motion and present a method that corrects noise-induced variance. This problem only becomes visible with long time windows and high noise levels.

\section{Our Approach}

Our method operates on the Image of Warped Events~\ref{eq:accumulate}, before calculating the image contrast~\ref{eq:variance}. It applies a multiplicative correction that solely depends on the motion candidate $\boldsymbol{\theta}$ and the width and height of the sensor. We first describe the method for one-dimensional sensors~\ref{sec:1d}. We then expand it to two-dimensional sensors~\ref{sec:2d}.

\begin{figure*}[h]
  \centering
  \includegraphics[width=6.8in]{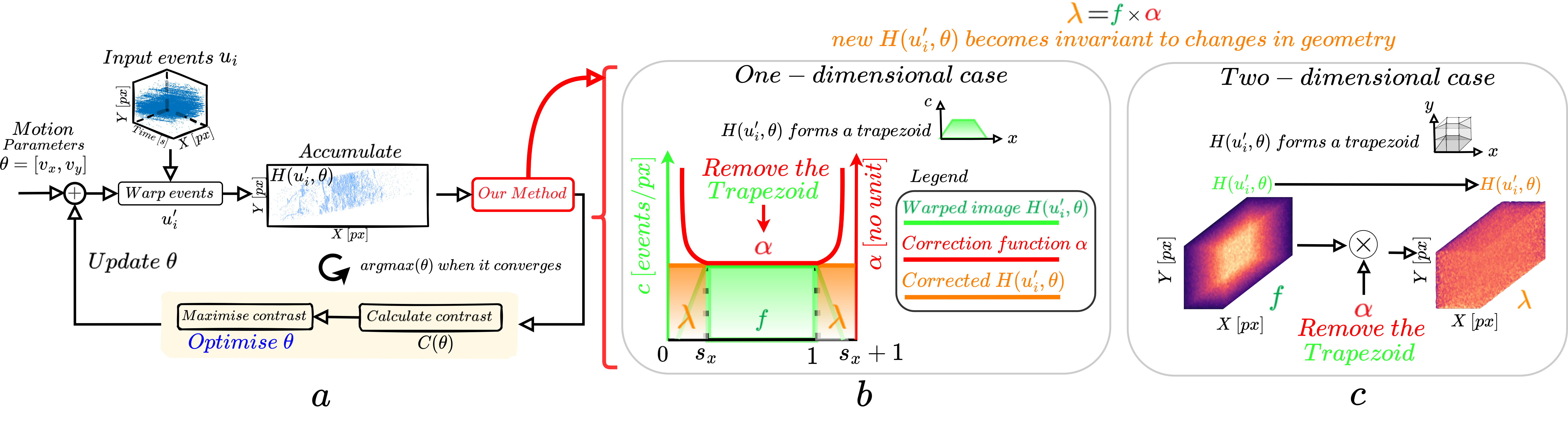}  \caption{A detailed overview of the proposed approach. \textbf{a:} The process of CMax including our method. \textbf{b:} The One-dimensional case shows how the trapezoid is formed after accumulating the warped event with the correction function $\alpha$ that removes it. \textbf{c:} The Two-dimensional case takes into account both pixel dimensions and removes the trapezoid, producing a new warped image that is invariant to geometry and capable of removing the multiple extrema from the loss landscape.}
   \label{fig:fullmethod}
\end{figure*}

\subsection{One-dimensional case}\label{sec:1d}

Let us consider an event sensor with a single row of pixels that generates random, uniformly distributed noise events with overall rate $\rho$ (in events per second). These events can be seen as a dense point cloud in two-dimensional space $\{x, t \}$. We approximate this point cloud with a "solid" rectangle in $\{x, t \}$ that spans the width of the sensor. Under this approximation, warping the events is equivalent to shearing the rectangle and results in a parallelogram in $\{x, t \}$. The transformation shifts the top of the rectangle by $-v \delta$, where $v$ is the candidate speed and $\delta$ is the considered time window.

We denote by $f$ the \textit{Line of Warped Events} (the one-dimensional equivalent of the Image of Warped Events). Its values $f(x)$ are given by the height of the parallelogram at $x$, hence their plot is a trapezoid without a base~(\ref{fig:fullmethod}). The "contrast" in the Line of Warped Events can be estimated with the variance of $f$ over the interval $[0, w - v \delta]$ (for $v \le 0$), denoted $\text{var}_f$. Importantly, we are not calculating the variance of a random variable but simply considering a continuous extension of the formula for the variance of a collection of samples. This is similar to the difference between the mean of a random variable and the mean of a function.

\begin{equation}
    \text{var}_f(v) = \frac{1}{w - v \delta}\int\limits_{0}^{w - v\delta}(f(x)-\overline{f})^2dx \quad \text{if}\  v \le 0
    \label{eq:variancestep1}
\end{equation}

$\overline{f}$ is the mean of f over the interval $[0, w - v \delta]$.

\begin{equation}
    \overline{f}(v) = \frac{1}{w - v \delta}\int\limits_{0}^{w - v\delta}f(x)dx \quad \text{if}\  v \le 0
    \label{eq:mean}
\end{equation}

For $v \ge 0$, one must consider the interval $[-v \delta, w]$ and divide by $w + v \delta$.

$\text{var}_f$ is zero if and only if $f$ is constant. That is the case for $v = 0$, however, all other candidate speeds result in a non-zero variance. This change in variance, which is created by sheared noise, causes the problem described in figure~\ref{fig:motivationproblems}. To show that our simple continuous model is sufficient to describe the observations, we calculate below an explicit expression of $\text{var}_f$ as a function of $v$.

Without loss of generality, we can restrict the problem to $v < 0$ and introduce unit-less variables to simplify the expressions.

\begin{itemize}
\item Normalised pixel position $p = \frac{x}{w}$
\item Normalised shear $s = \frac{v \delta}{w}$
\item Normalised event count $c = \frac{\delta}{\rho}$
\end{itemize}

\begin{figure}[ht]
  \centering
  \includegraphics[width=3in]{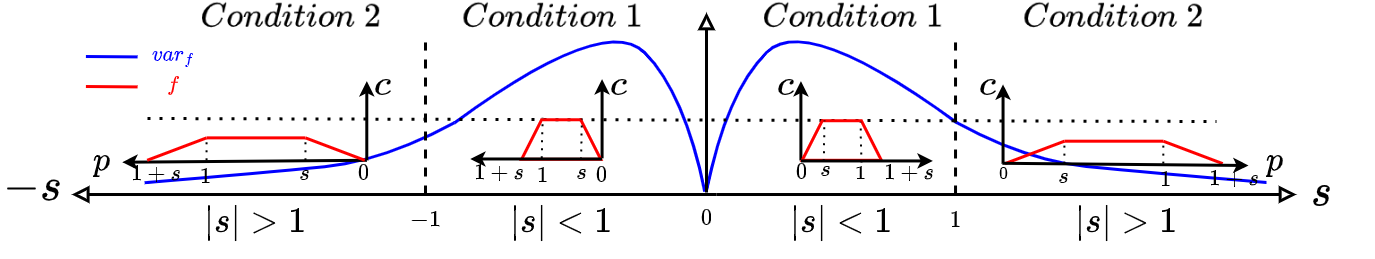}
  \caption{Warping dense noise is similar to shearing a "solid" rectangle. The operation turns the rectangle into a parallelogram. The height of the parallelogram as a function of the pixel position describes the top of a trapezoid, shown here in red. When $s$ is larger than one (the time window times the speed is larger than the sensor width), the curve's maximum value decreases. The variance of the red curve as a function of the shear, in blue, follows a surprisingly complex pattern, with a maximum at $s = \frac{1}{2}$.}
   \label{fig:geometry1d}
\end{figure}

$f(p)$ is a piecewise linear function made of three segments, with two slightly different expressions for $s \le 1$ and $s \ge 1$ (figure~\ref{fig:geometry1d}). These two expressions correspond (respectively) to shears smaller than the sensor width and shear larger than the sensor width.

For $s \le 1$, $f$ is defined on $[0, 1 + s]$ by:

\begin{equation}
 f(p) = \begin{cases}
    \frac{cp}{s} & 0 \le p \le s \\
    c & s \le p \le 1  \\
    c \left(1-\frac{p-1}{s}\right) & 1 \le p \le 1 + s \\
   \end{cases} \label{eq:1df1_pw}
\end{equation}

Applying the formulas for the mean and variance given for $s \le 1$:

\begin{equation}
    \overline{f} = \frac{c}{s+1} \quad \text{and}\quad \text{var}_f(s)= c^{2}\frac{ s \cdot \left(2 - s\right)}{3 \left(s + 1\right)^{2}}\label{eq:varianceoutputshear}
\end{equation}

\begin{figure}[ht]
  \hspace*{-1cm}
  \centering
  \includegraphics[width=4in]{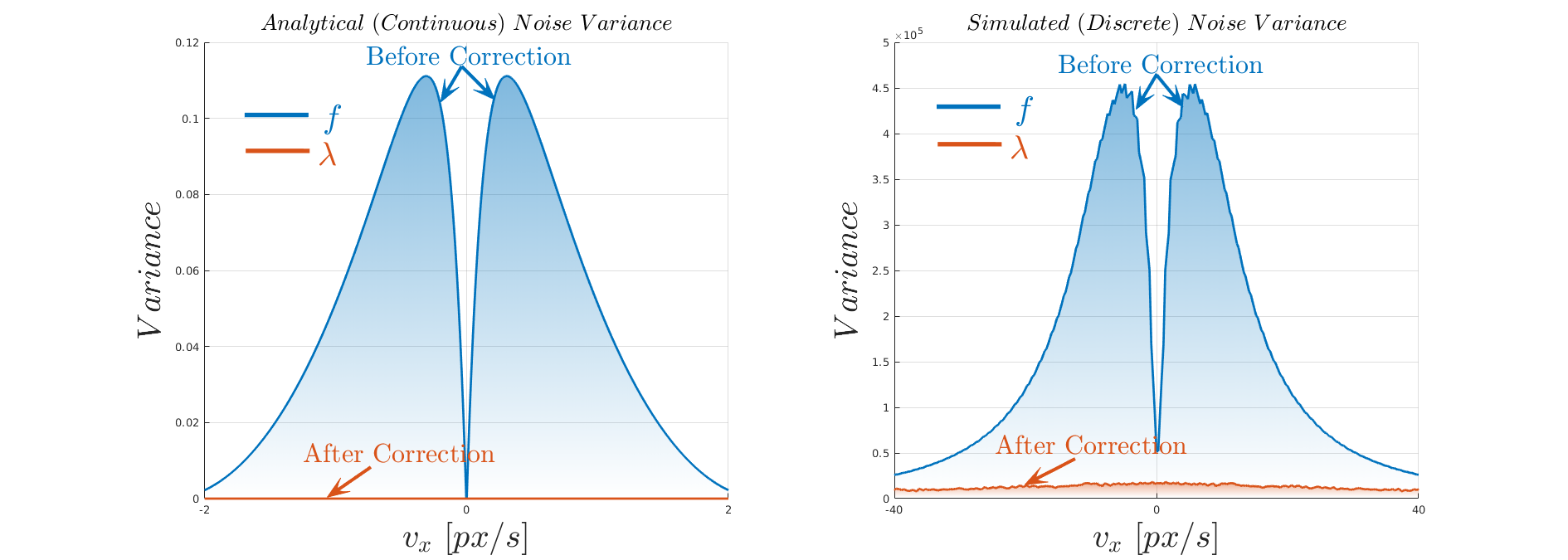}
  \caption{Plots of the variance of $f$ and $\lambda$ as a function of the candidate speed, using the analytical formula (\ref{eq:varianceoutputshear}) (left) and a discrete simulation (right).}
   \label{fig:1Dresults}
\end{figure}

The equations for $s \ge 1$ are given in the supplementary materials (Section II). Plotting $\text{var}_f$ yields a figure that is very similar to a section of the velocity landscape obtained from real event sensor data (\ref{fig:1Dresults}). The function $\text{var}_f$ admits a maximum at $s = \frac{1}{2}$ when the velocity multiplied by the time window equals half the sensor width.

We want $\text{var}_f$ to be zero since $f$ represents the Line of Warped Events for uniform noise. We thus introduce $\alpha$, a multiplicative correction function for the non-constant segments of $f$ (the slopes of the trapezoid). $\alpha$ is defined on $[0, 1 + s]$ by:

\begin{equation}
   \alpha(p) = \begin{cases}
    \frac{s}{p} & 0 \le p \le s \\
    1 & s \le p \le 1  \\
    \frac{s}{s + 1 - p} & 1 \le p \le 1 + s \\
   \end{cases} \label{eq:1dcorrectionfunction}
\end{equation}

Since $\alpha$ is a multiplicative correction rather than an additive one, it does not depend on $c$ (and, by extension, $\rho$). In other words, the correction can be applied with no prior knowledge of the noise density. The corrected counterpart of $f$, denoted $\lambda$, is defined by:

\begin{equation}
    \lambda(p) = \alpha(p) \cdot f(p)
    \label{eq:lambda}
\end{equation}

By construction, $\lambda$ is equal to the constant $c$ on $[0, 1 + s]$, and its variance is trivially zero. Figure \ref{fig:1Dresults} shows a plot of the variance expression and demonstrates the correction on simulated discrete noise events.

\subsection{Two-dimensional case}\label{sec:2d}

In this section, we extend our model to two-dimensional sensors. Let us consider an event sensor with $w\times h$ pixels that generate random, uniformly distributed noise events with the overall rate $\rho$ (in events per second), a speed candidate $\mathbf{\theta} = \left[v_x, v_y \right]$, and a time window $\delta$. We model the noise events as a "solid" rectangular cuboid in $\{x, y, t\}$ space.

We define the following normalised variables to express the variance in two dimensions:

\begin{itemize}
\item Normalised x pixel position $p_x = \frac{x}{w}$
\item Normalised y pixel position $p_y = \frac{y}{h}$
\item Normalised shear alongside the x-axis $s_x = \frac{v_x \delta}{w}$
\item Normalised shear alongside the y-axis $s_y = \frac{v_y \delta}{h}$
\item Normalised event count $c = \frac{\delta}{\rho}$
\end{itemize}

\begin{figure}[ht]
  \centering
  \includegraphics[width=3in]{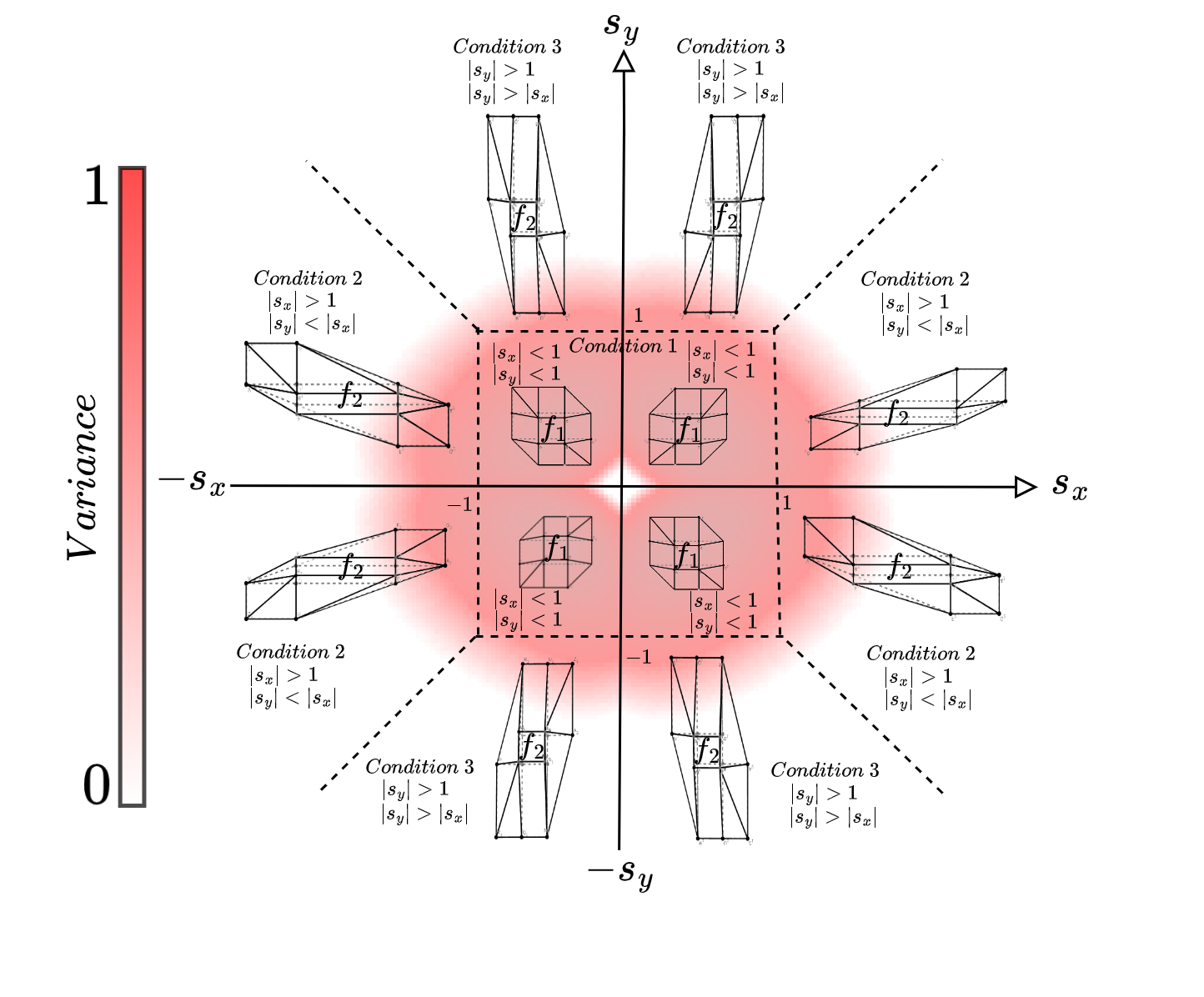}
  \caption{An illustration of the height of the sheared rectangular cuboid (\ref{eq:accumulate}) as a function of $s_x$ and $s_y$ (black geometric figures) and the corresponding variance (red and white background). The problem exhibits several symmetries and can be solved by considering only two sets of conditions (condition 2 and condition 3 are symmetrical about the axis $y = x$).}
   \label{fig:geometry2d}
\end{figure}

Shearing the rectangular cuboid in two directions yields a parallelepiped. Unlike the one-dimensional case, in which the height of the sheared geometry was another well-known figure, the height of the parallelepiped at every point of the $\{x, y\}$ plane forms a "generic" polyhedron with 7 faces without counting the base (figure~\ref{fig:geometry2d}).

\begin{figure}[h!]
  \centering
  \includegraphics[width=3.5in]{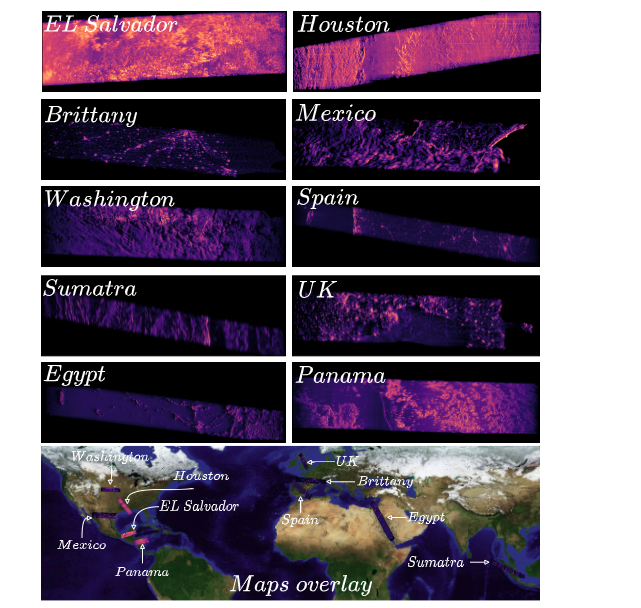}
  \caption{Motion-compensated maps. Top: The $H(\boldsymbol{u'},\boldsymbol{\theta})$ image using the motion parameters $\boldsymbol{\theta}$ estimated by our approach. Bottom: An overlay of motion-compensated maps over the earth map. The overlay was performed manually and the scale was a bit exaggerated to better see the motion-compensated maps.}
  \label{fig:finalmaps}
\end{figure}

Another complication in two dimensions is the shape of the integration domain. While it would be tempting to use a rectangular integration domain, some parts of the scenes are never in the field of view (specifically, two triangular regions of combined area $s_x s_y$, as shown in the Houston recording in figure~\ref{fig:finalmaps}). We take this into account in the integration calculation and divide by $(1+s_x)(1+s_y)-s_xs_y$ in the mean and variance formulas. 

For $s_x \le 1$ and $s_y \le 1$, the height function $f$ is defined by:

\begin{equation}
f(p_x, p_y) = \begin{cases}
    \frac{cp_y}{s_y} & p_x \le s_x \land p_y \le \frac{s_y}{s_xp_x} \\
    \frac{cp_x}{s_x} & p_x \le \frac{s_x}{s_yp_y} \land p_y \le s_y  \\
    \frac{cp_y}{s_y} & s_x \le p_x < 1 \land p_y \le s_y \\
    \frac{cp_x}{s_x} & p_x \le s_x \\ & \land\ s_y \le p_y \le 1 \\
    c & s_x \le p_x \le 1 \\ & \land\ s_y \le p_y \le 1 \\
    c \left(\frac{1 - p_x}{s_x}+\frac{p_y}{s_y} \right) & 1 \le p_x \le 1+s_x \\ & \land\ (p_x-1)\frac{s_y}{s_x} \le p_y \le s_y
   \end{cases} \label{eq:f_piecewise}
\end{equation}

Calculating the mean and variance of $f$ yields:

\begin{equation}
    \overline{f}=\frac{c}{s_x + s_y + 1}\label{eq:2dmeanv2}
\end{equation}

\begin{equation}
    \text{var}_f(s_x, s_y)= c^{2}\frac{\left(s_{x}^{2} s_{y} + s_{x} s_{y}^{2} - 3 s_{x} s_{y} + q_x + q_y \right)}{6 \left(s_{x} + s_{y} + 1\right)^{2}}\label{eq:2dvariancev2}
\end{equation}

where $q_x = - 2 s_x^2 + 4 s_x$ and $q_y = -2 s_{y}^{2} + 4 s_{y}$.

\begin{figure}[ht]
  \hspace*{-1.5cm}
  \centering
  \includegraphics[width=3.8in]{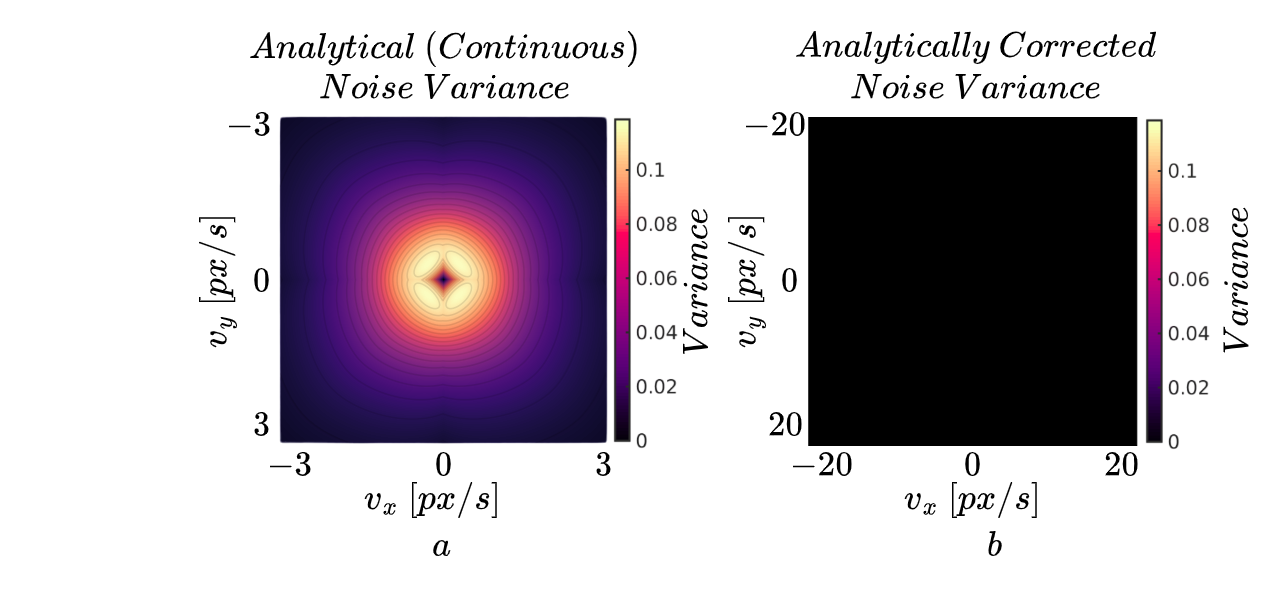}
  \caption{The left graph~(\textbf{a}) is a plot of the analytical formula for the variance as a function of the candidate speed. The right graph~(\textbf{b}) is a plot of the corrected variance calculated on discrete simulated noise, and is zero as intended.}
   \label{fig:2Dvaranalytics}
\end{figure}

Figure~\ref{fig:2Dvaranalytics} shows a plot of the variance as a function of $s_x$ and $s_y$. Despite its simplicity, our model predicts quite well the "ring" that we observe on real sensor data. Similarly to the one-dimensional case, we introduce a multiplicative correction function $\alpha$ to "flatten" $f$ and ensure that the variance of the corrected height function is zero. For $s_x \le 1$ and $s_y \le 1$, $\alpha$ is defined by:

\begin{equation}
    \alpha(p_x,p_y) = \begin{cases}
    \frac{s_y}{p_y} & p_x \le s_x \land p_y \le \frac{s_y}{s_xp_x} \\
    \frac{s_x}{p_x} & p_x \le \frac{s_x}{s_yp_y} \land p_y \le s_y  \\
    \frac{s_y}{p_y} & s_x \le p_x \le 1 \land p_y \le s_y \\
    \frac{s_x}{p_x} & p_x \le s_x \land s_y \le p_y \le 1 \\
    1 & s_x \le p_x \le 1 \\ & \land\ s_y \le p_y \le 1 \\
    \frac{s_xs_y}{p_ys_x-s_yp_x+s_y} & 1 \le p_x \le 1+s_x \\ & \land\ (p_x-1)\frac{s_y}{s_x} \le p_y \le s_y
   \end{cases} \label{eq:alphacorrection}
\end{equation}

The equations for $s_x \geq 1$ and $s_y \geq 1$ are given in the supplementary materials (Section III). To obtain a corrected warped image from real event data, we can directly apply the correction $\alpha$ to the pixels of the accumulated image (\ref{eq:accumulate}), given only the candidate warp speed $[v_x, v_y]$ and the dimensions of the sensor. This minimises the contribution of uniform noise to the variance in the image and produces a new accumulated frame denoted as $H(\boldsymbol{u'_i},\boldsymbol{\theta})_{v}$, as described in (\ref{eq:finalcorrection}). Finally, we can calculate the variance using (\ref{eq:variance}) on the corrected image to solve the problem.

\begin{equation}
    H(\boldsymbol{u'_i},\boldsymbol{\theta})_{v} = H(\boldsymbol{u'_i},\boldsymbol{\theta})\times\alpha \label{eq:finalcorrection}
\end{equation}

We employed analytical integration techniques based on motion and geometry to ensure that the variance is only high around correct $\boldsymbol{\theta}$. By extending our approach to 2D, we can now apply this corrective technique to real-world data. In the next section, we demonstrate how this method can be used to generate motion-compensated images from data acquired from the ISS.

\section{Experiments}\label{sec:experiments}

To show how our method successfully generalises to real event data, we now apply this technique to data in which noise heavily dominates the structure of the events. We evaluate our approach using data captured directly with an event camera on the ISS. The dataset comprises recordings that vary from 30 to 180 seconds and were captured under a diverse set of conditions. These include day/night recordings, different locations on earth, and varying weather conditions. The recordings contain an average of 7 million events per recording - See supplementary materials (Section I). The dataset does not come with an associated evaluation protocol and we,  therefore, have defined an evaluation protocol based on RMS and a new metric called the Rate of Convergence (RoC). The RoC shows the rate of success of the optimisation algorithm in converging to the correct motion values. For simplicity, we used the Nelder-Mead optimisation (NMO) algorithm to search for the correct motion parameters and calculated the RoC by initialising the optimisation algorithm at every single point between -30$px/s$ and 30$px/s$ and then calculating the overall percentage of how many times it successfully converges. The ground truth $\boldsymbol{\theta}$ was manually found for each recording.

\subsection{Qualitative Results}\label{sec:quality}
\begin{figure*}[t]
\renewcommand*{\arraystretch}{0}\centering
\begin{tabular}{{M{0cm}M{3.5cm}M{3.5cm}M{0cm}M{3.5cm}M{3.5cm}}}
 &   \ \  \textbf{CMax\cite{gallego_unifying_2018}} &   \textbf{Ours} & &\ \  \textbf{CMax\cite{gallego_unifying_2018}} &  \textbf{Ours}\\
 \large  \rotatebox{90}{EL Salvador} & \includegraphics[width=1.6in]{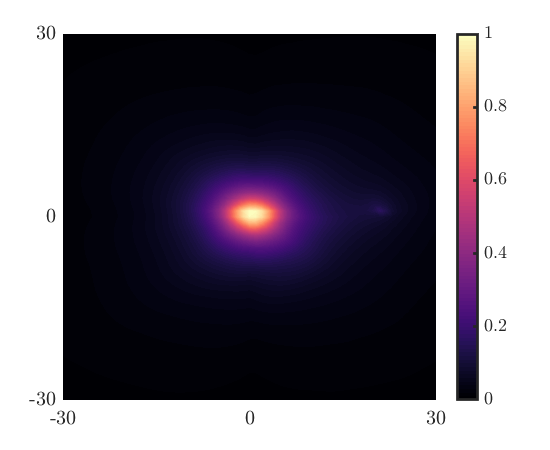} & \includegraphics[width=1.6in]{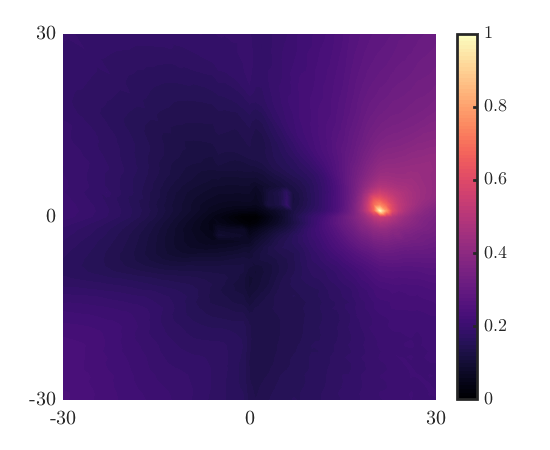} &
  \large  \rotatebox{90}{Houston} &\includegraphics[width=1.6in]{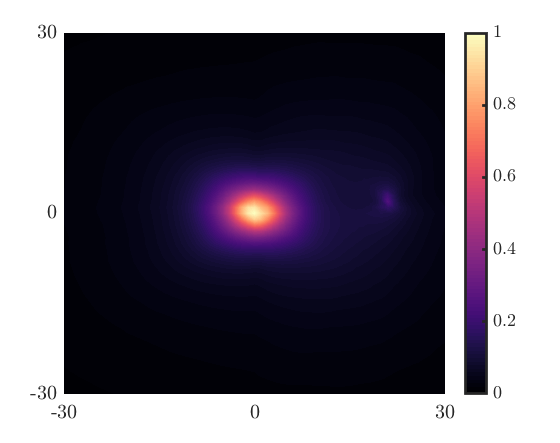} &
  \includegraphics[width=1.6in]{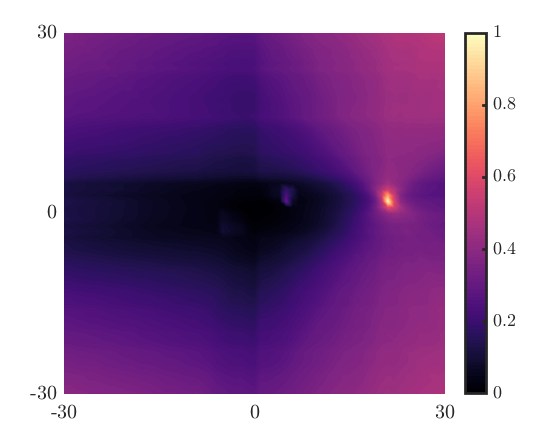}\\
  \large\rotatebox{90}{Brittany} &\includegraphics[width=1.6in]{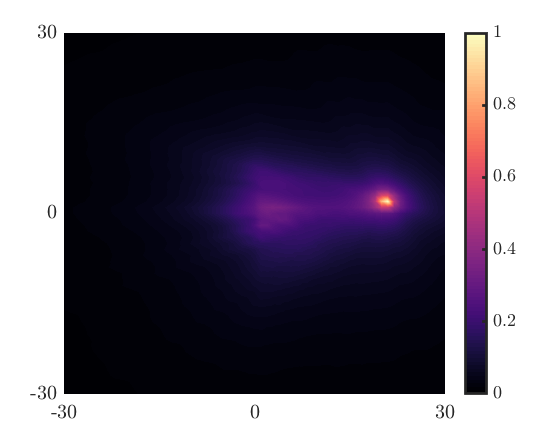} &
  \includegraphics[width=1.6in]{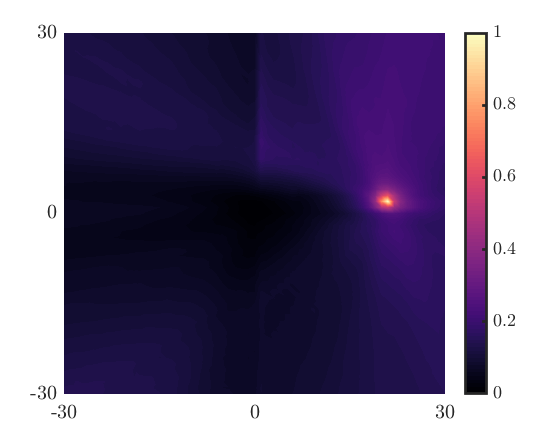} &
  \large\rotatebox{90}{Mexico} &\includegraphics[width=1.6in]{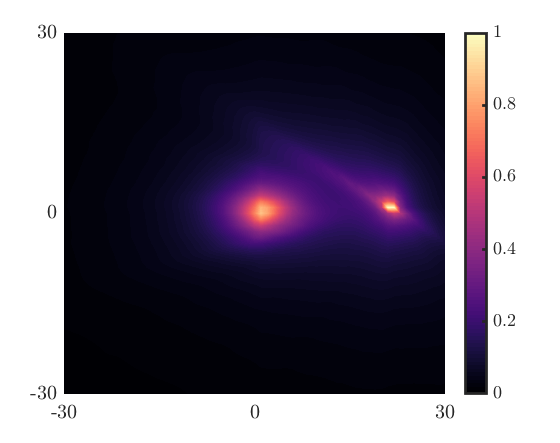} &
  \includegraphics[width=1.6in]{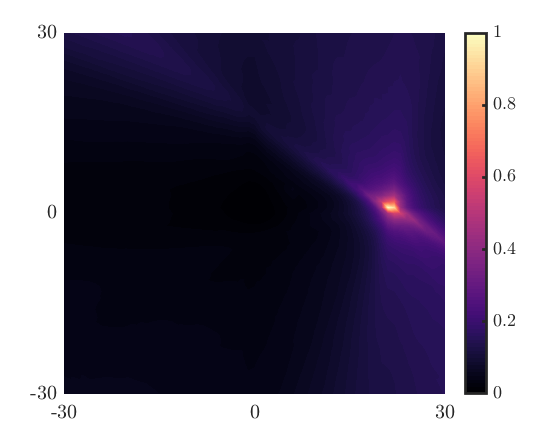}\\
  \large\rotatebox{90}{Washington} & \includegraphics[width=1.6in]{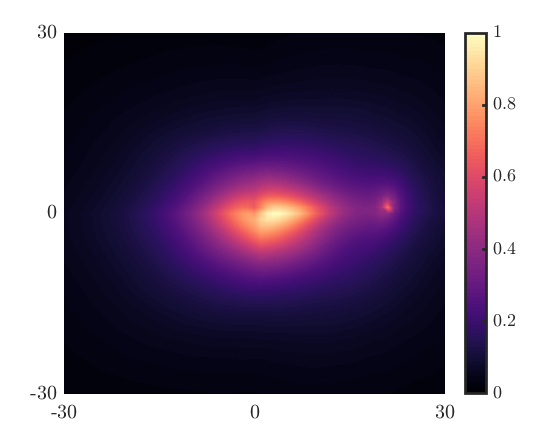} & 
  \includegraphics[width=1.6in]{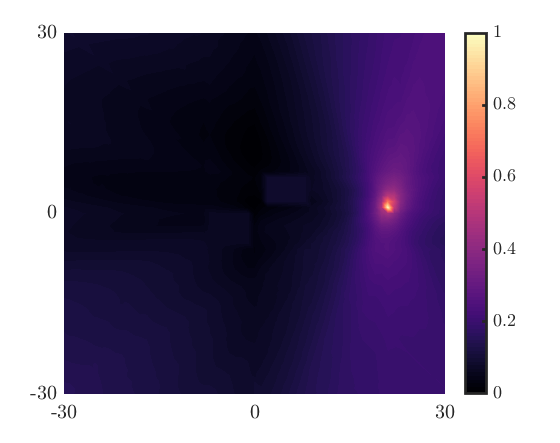} & 
  \large  \rotatebox{90}{Spain} &\includegraphics[width=1.6in]{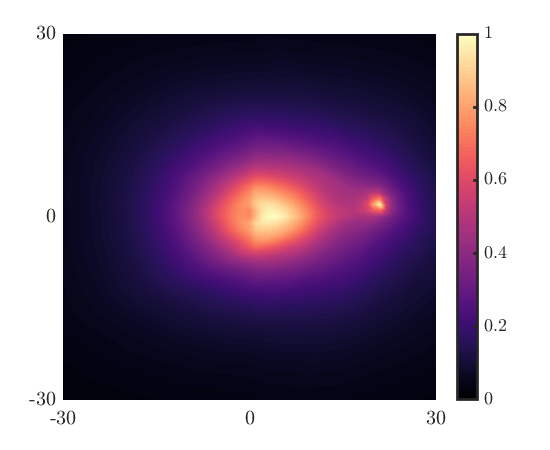} & 
  \includegraphics[width=1.6in]{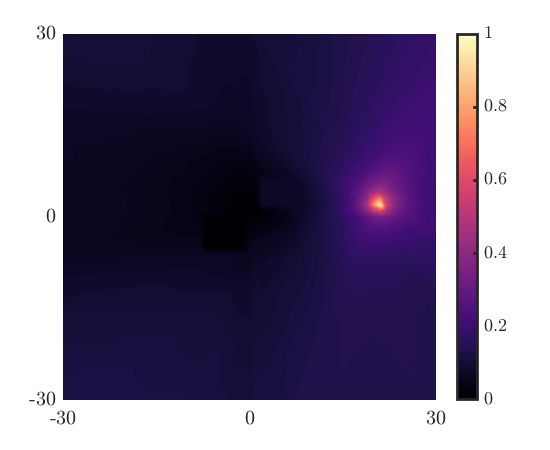}\\
  \large\rotatebox{90}{Sumatra}&\includegraphics[width=1.6in]{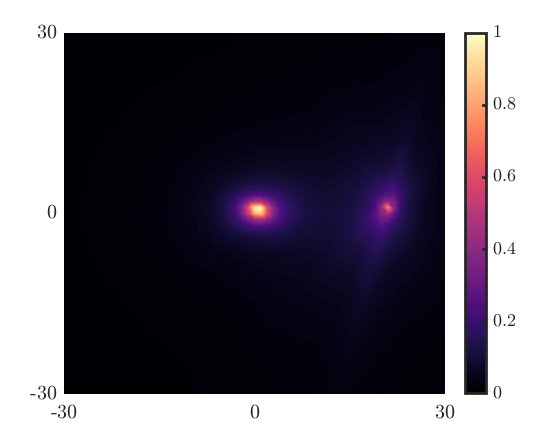} &
  \includegraphics[width=1.6in]{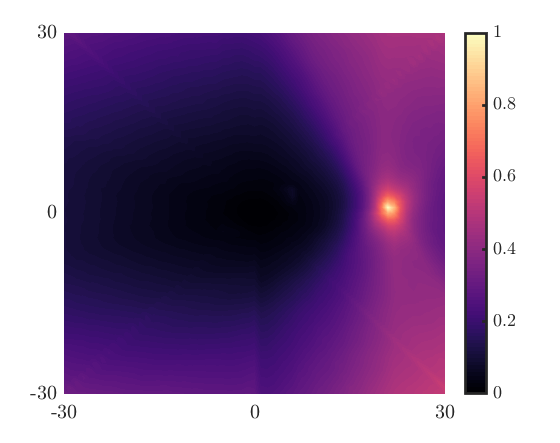}  &
  \large  \rotatebox{90}{UK} &\includegraphics[width=1.6in]{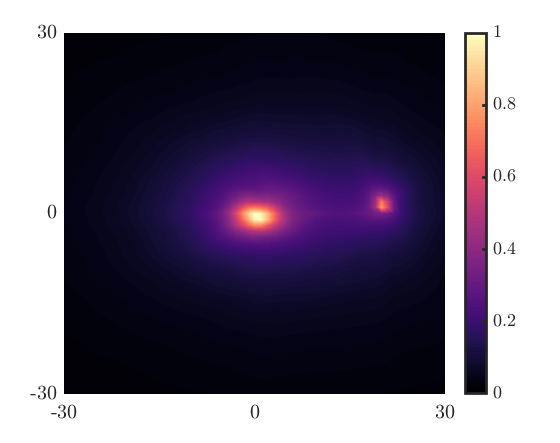}&
  \includegraphics[width=1.6in]{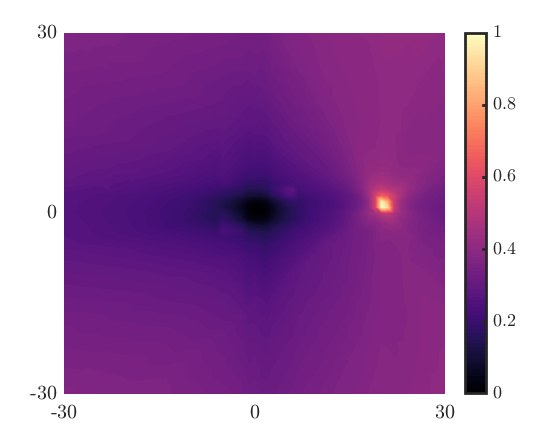}\\
  \large\rotatebox{90}{Egypt}&\includegraphics[width=1.6in]{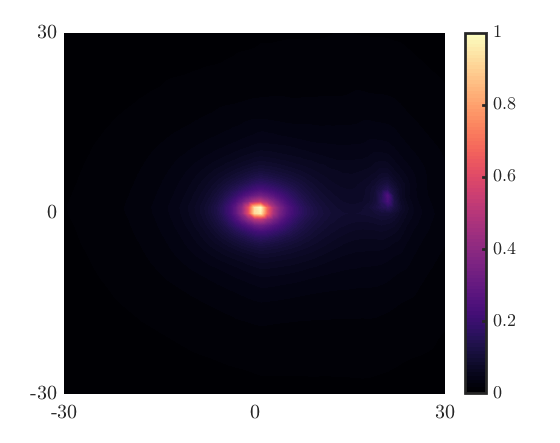}&
  \includegraphics[width=1.6in]{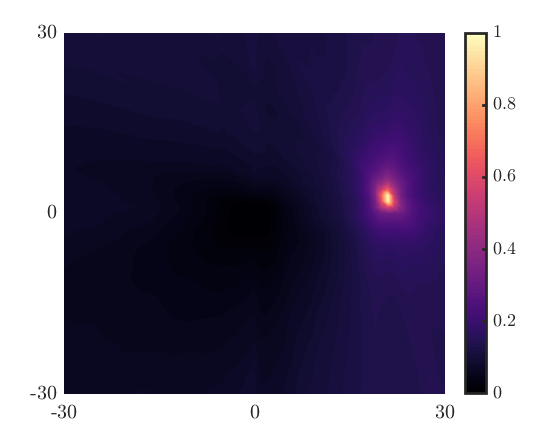}&
  \large  \rotatebox{90}{Panama} &\includegraphics[width=1.6in]{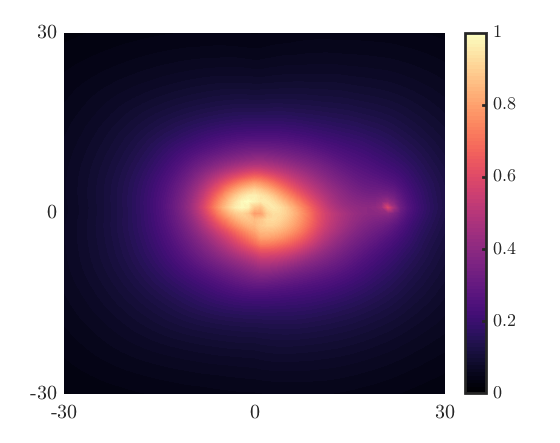}&
  \includegraphics[width=1.6in]{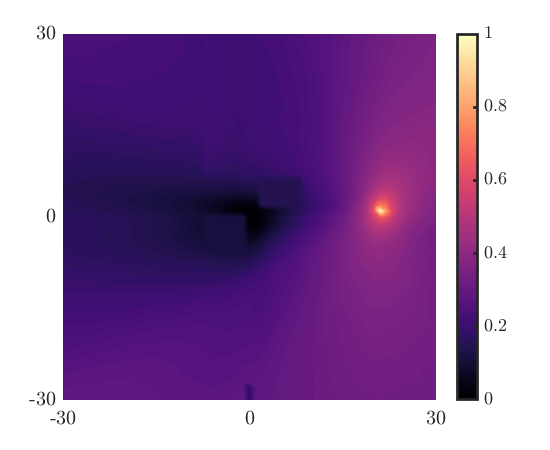}\\
\end{tabular}
\caption{Comparing the loss landscape produced by CMax \cite{gallego_unifying_2018} and our approach. Our approach shows a single solution corresponding to the correct motion $\boldsymbol{\theta}$, while CMax shows multiple maxima in each case. X-axis represents $v_x \ [px/s]$ and Y-axis represents $v_y \ [px/s]$ as a function of the variance (\ref{eq:variance}), normalised between 0-1.}
\label{tb:qualitativeresults}
\end{figure*}

Figure \ref{tb:qualitativeresults} shows qualitative results using our proposed method compared with standard CMax \cite{gallego_unifying_2018}. Our approach always produces a single solution around the correct motion parameters, whereas CMax shows multiple extrema in each case where the global maximum is much more prominent than the correct local maximum. A single correct motion solution indicates that the geometry of the IWE and the density of the events are no longer affecting the variance calculation. Our method also leads to better motion-compensated and sharp maps, which can be used for matching with existing satellite images and other orbital-related applications as in Figure \ref{fig:finalmaps}.

\subsection{Quantitative Results}\label{sec:quatity}

In this section, we thoroughly examine the effectiveness of our proposed approach by comparing its performance with the existing CMax algorithm. We evaluate the results in terms of RMS and RoC. As shown in Table \ref{tab:evaluation}, our approach significantly outperforms CMax in every case.

\begin{table}[h]
\tiny
\centering
\caption{Quantitative results. Final results of the ISS data comparing CMax \cite{gallego_unifying_2018} with our approach with RMS error and Rate of Convergence (RoC) as evaluation metrics.}
\label{tab:evaluation}
\resizebox{0.4\textwidth}{!}{%
\begin{tabular}{lcccc}
\hline
\multicolumn{1}{c}{\multirow{2}{*}{}} & \multicolumn{2}{c}{CMax \cite{gallego_unifying_2018}} & \multicolumn{2}{c}{Ours} \\ \cline{2-5}
\multicolumn{1}{c}{} & $RMS$ & $RoC\%$ & $RMS$ & $RoC\%$ \\ \hline\hline
EL Salvador & 14.47 & 2.55 & \textbf{0.61} & \textbf{75.57} \\ \cline{1-1}
Houston & 13.74 & 2.62 & \textbf{0.55} & \textbf{81.48} \\ \cline{1-1}
Brittany & 0.08 & 83.13 & \textbf{0.01} & \textbf{83.57} \\ \cline{1-1}
Mexico & 14.13 & 1.16 & \textbf{0.09} & \textbf{80.50} \\ \cline{1-1}
Washington & 14.19 & 2.87 & \textbf{0.11} & \textbf{74.10} \\ \cline{1-1}
Spain & 13.77 & 2.45 & \textbf{0.14} & \textbf{80.82} \\ \cline{1-1}
Sumatra & 13.41 & 1.62 & \textbf{0.22} & \textbf{81.60} \\ \cline{1-1}
UK & 12.84 & 2.02 & \textbf{0.28} & \textbf{82.89} \\ \cline{1-1}
Egypt & 13.53 & 1.95 & \textbf{0.01} & \textbf{76.51} \\ \cline{1-1}
Panama & 14.50 & 2.72 & \textbf{0.04} & \textbf{70.61} \\ \hline
\end{tabular}%
}
\end{table}

The low RoC in CMax can be attributed to the presence of the global maximum, which often causes the NMO to be stuck at the global maximum and produce a high RMS error. Our method guarantees a single local maximum, which enables the NMO to converge successfully, resulting in a higher RoC.

Furthermore, the RoC of CMax is usually low since it only succeeds when the NMO is initialised close to the true motion parameters. In contrast, our approach is more robust and works effectively even when the global maximum is not dominant, as demonstrated in the Brittany case. Here, both methods converged since the recording had less noise compared to the rest of the data. This demonstrates the flexibility and general applicability of our approach to handling events with various density levels.

\section{Conclusion}
In this paper, we present an analytical solution to the noise-intolerance and the multiple extrema problems of the CMax framework. Our solution is purely based on geometrical principles and the physical properties of the events. First, we analysed these problems in 1D and 2D spaces by considering the events as a solid rectangle in 1D, and as a solid rectangular cuboid in 2D to demonstrate the influence of the change in geometries on the variance calculation. We then demonstrated how our analytical solution makes CMax invariant to the changes in the geometry and avoids having high contrast around the wrong motion parameters, without using any prior of the camera motion and regardless of the density of the events. The experimental results demonstrate the superior performance of our method compared to the state-of-the-art CMax when used on extremely noisy data. \\

\textbf{Acknowledgement}. This work was supported by AFOSR grant FA2386-23-1-4005 and USAFA grant FA7000-20-2-0009.

{\small
\bibliographystyle{ieee_fullname}
\bibliography{egbib}
}

\end{document}


\maketitle

In this document, we describe our novel dataset and provide an additional detailed explanation of our mathematical approach in the paper "Density Invariant Contrast Maximization for Neuromorphic Earth Observations".

\section{ISS Dataset}
\begin{figure}[h]
  \centering
  \includegraphics[width=7in]{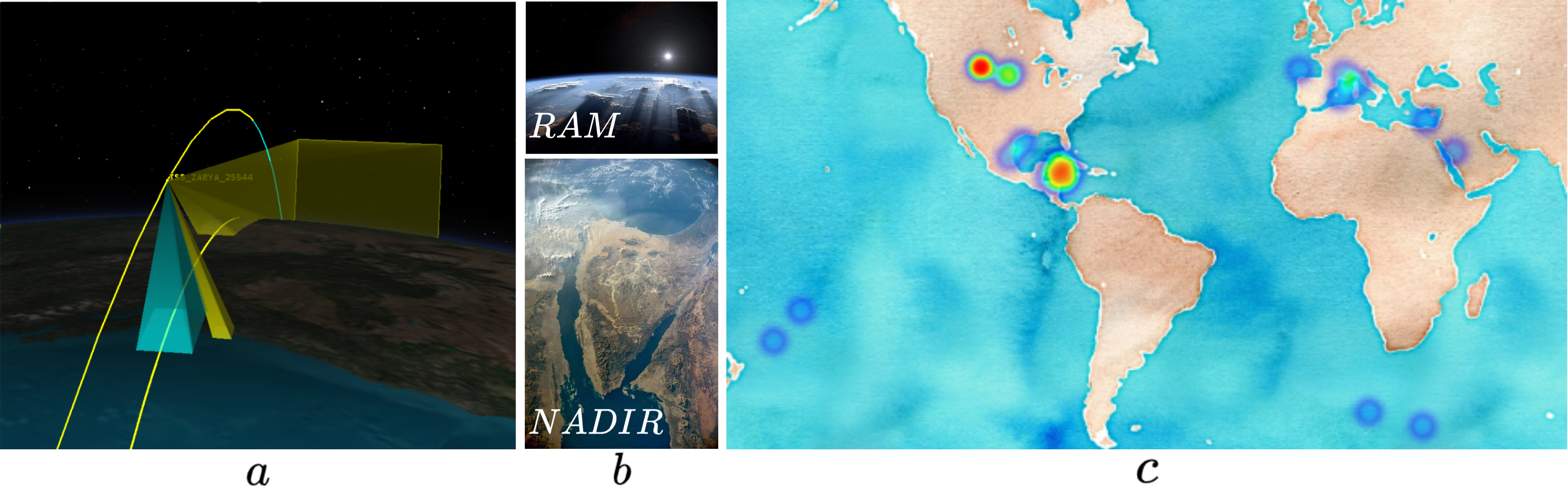}
  \caption{The FOV of the event camera on the ISS. \textbf{a:} The yellow shade represents the FOV of two event cameras, in this work, we use the event camera pointed toward earth or NADIR. \textbf{b:} An example of what a normal camera sees through the RAM and NADIR FOV. \textbf{c:} A heatmap of the recording locations which are selected for this paper.}\label{fig:issschematic}
\end{figure}

The ISS-based event camera setup consists of two DAVIS cameras, each with a resolution of 240x180 pixels, located at the Columbus module. One camera referred to as the "Ram camera," points forward toward the Earth's limb in the direction of the ISS's travel. The other camera, known as the "Nadir camera," points down towards the Earth at a 20-degree angle to starboard to observe lightning and the Earth's surface directly from above the upper atmosphere. The camera field-of-view (FOV) is depicted in Figure \ref{fig:issschematic}(a) with yellow shades representing both the Nadir and Ram FOV and the blue shade representing the ISS centre FOV at its midpoint. Figure \ref{fig:issschematic}(b) illustrates the FOV for both cameras from the Nadir and Ram perspectives, showcasing the Ram's zooming ability and Nadir's translation motion. The Ram is primarily used to observe sprites and lightning from different angles, while the Nadir captures lightning and Earth's surface. We carefully selected ten recordings from the Nadir camera dataset, considering a variety of conditions such as weather, location, time, and noise. Table \ref{tab:ISScharacteristics2} provides additional details about each recording. We focused on recordings that exhibit rich textures and clear outlines of the Earth, disregarding recordings over the ocean, as they only generate noise without features or clear outlines, rendering them unusable.

\begin{longtable}[c]{lccccc}
\caption{General characteristics of the ISS event dataset.}
\label{tab:ISScharacteristics2}\\
\hline
\textbf{Geo Location} & \textbf{Date (UTC)} & \textbf{$\Delta t \ (s)$} & \textbf{Location$^{\circ}$} & \textbf{\# Events}  & \textbf{Earth Side} \\ \hline
\endhead
El Salvador & 2022-01-21 20:58:34 & 60 & \href{https://maps.google.com/?q=15.459,-82.117&ll=15.459,--82.117&z=3}{15.45 -82.11}  & 11,550,402 & Day\\ \hline
  Panama & 2022-01-24 20:12:11 & 30 & \href{https://maps.google.com/?q=16.818,-86.619&ll=16.818,-86.619&z=3}{16.81 -86.61}  & 13,876,281 & Day\\ \hline
  Brittany & 2022-01-25 21:21:18 & 60 & \href{https://maps.google.com/?q=44.418,5.334&ll=44.418,5.334&z=3}{44.41 5.33} & 841,024 & Night \\ \hline
  Mexico & 2022-01-25 20:58:52 & 90 & \href{https://maps.google.com/?q=23.168,-98.834&ll=23.168,-98.834&z=3}{23.16 -98.83} & 1,020,556  & Day \\ \hline
  Spain & 2023-01-13 03:17:57 & 30 & \href{https://maps.google.com/?q=-0.958,36.535&ll=-0.958,36.535&z=3}{-0.958 36.535} & 3,928,491  & Night\\ \hline
  Washington & 2022-02-01 20:15:58 & 30 & \href{https://maps.google.com/?q=43.797,-100.187&ll=43.797,-100.187&z=3}{43.79 -100.18 } & 3,280,126   & - \\ \hline
  Houston  & 2022-02-17 20:28:02 & 59 & \href{https://maps.google.com/?q=20.754,-84.596&ll=20.754,-84.596&z=3}{20.75 -84.59 } & 4,996,157  & Night \\ \hline
  Sumatra  & 2022-02-17 21:20:49 & 180 & \href{https://maps.google.com/?q=-48.63,13.163&ll=-48.63,13.163&z=3}{-48.63 13.163} & 1,073,675  & Night \\ \hline
  Egypt & 2022-02-03 17:28:03 & 180 & \href{https://maps.google.com/?q=32.942,30.969&ll=32.942,30.969&z=3}{32.94 30.96} &  6,426,732 & - \\ \hline
  United Kingdom & 2023-01-19 20:25:10 & 60 & \href{https://maps.google.com/?q=8.157,51.637&ll=8.157,51.637&z=3}{8.157 51.637} &  3,712,626  & - \\ \hline
\end{longtable}

\begin{figure*}[h]
  \centering
  \includegraphics[width=5.5in]{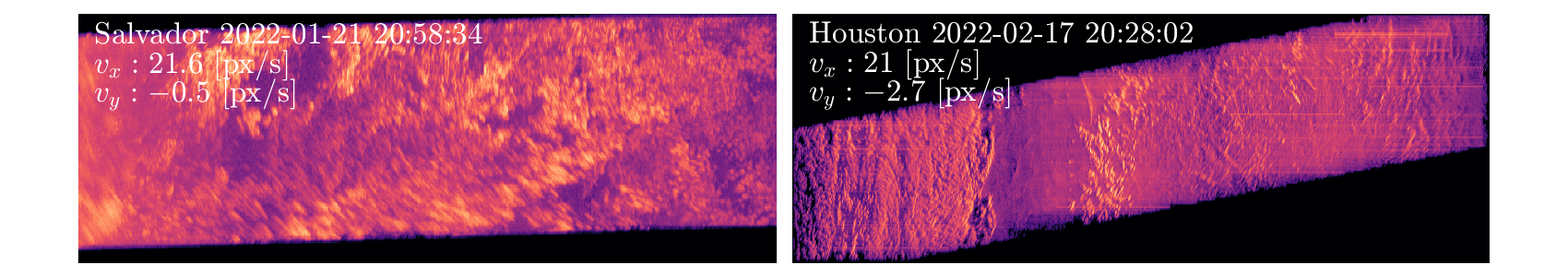}
  \includegraphics[width=5.5in]{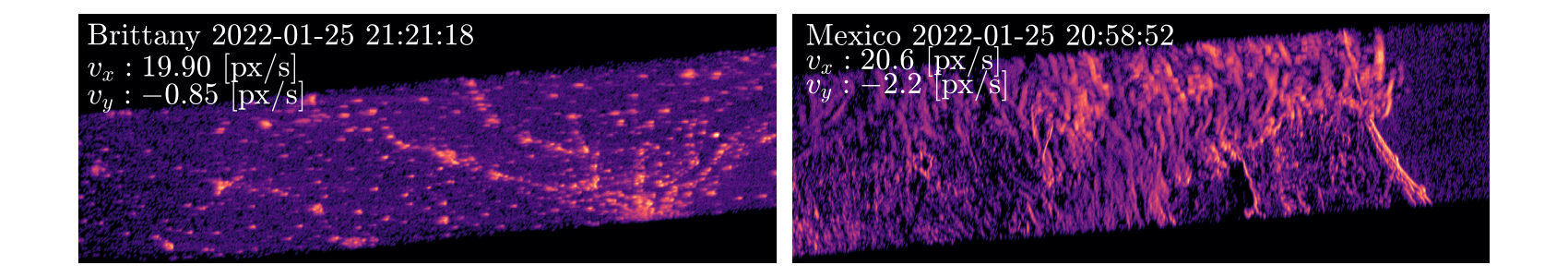}
  \includegraphics[width=5.5in]{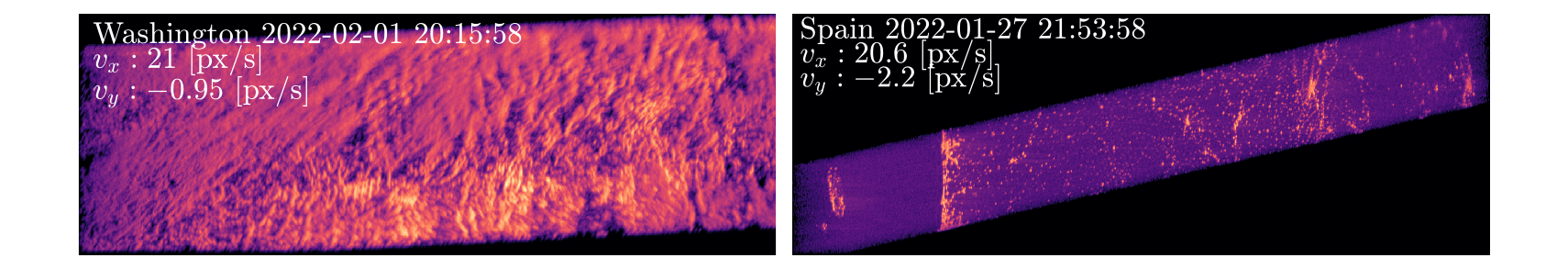}
  \includegraphics[width=5.5in]{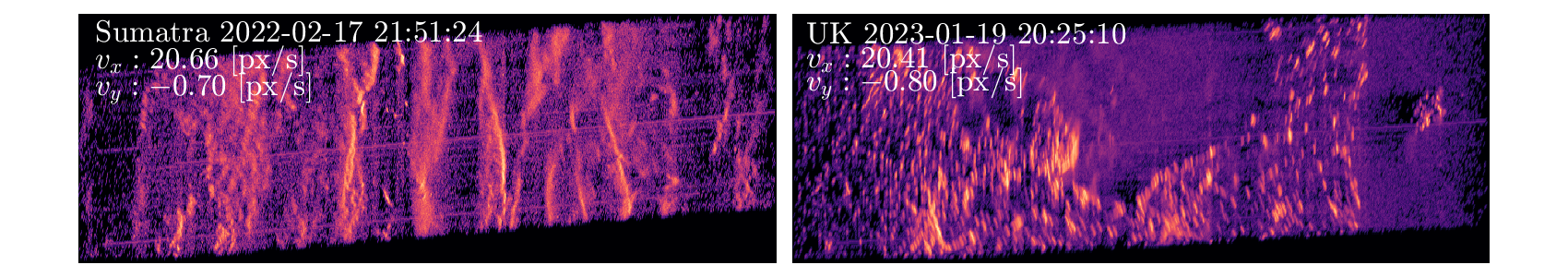}
  \includegraphics[width=5.5in]{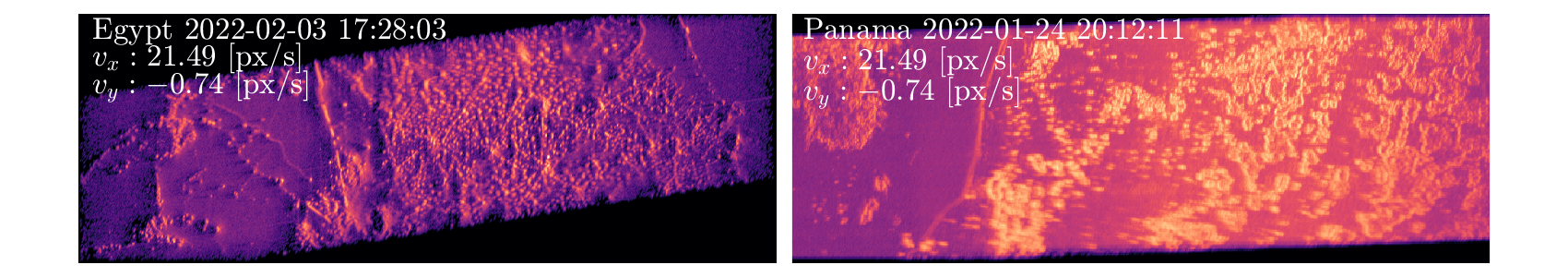}
  \caption{ISS motion-compensated maps.}
  \label{fig:motioncompensatedmaps}
\end{figure*}

Below are some snapshots of each data we used in this work (figure \ref{tab:sequencesdata}) with the output maps (figure \ref{fig:motioncompensatedmaps}) which are similar to the ones presented in the paper but with a larger size:

\begin{figure*}[h]
\centering
\begin{tabular}{cccccc}
\rotatebox{90}{\textbf{El Salvador}} & \includegraphics[width=1in]{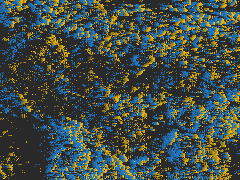} & \includegraphics[width=1in]{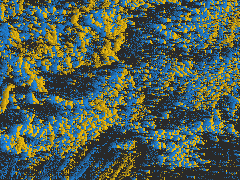} & \includegraphics[width=1in]{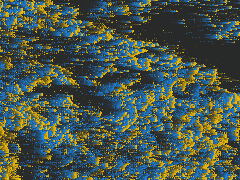} & \includegraphics[width=1in]{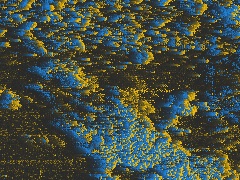} & \includegraphics[width=1in]{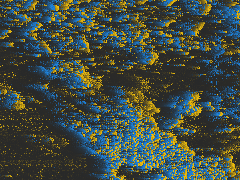}\\
\rotatebox{90}{\textbf{Houston}} & \includegraphics[width=1in]{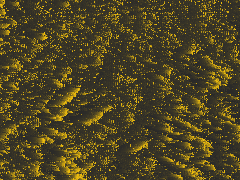} & \includegraphics[width=1in]{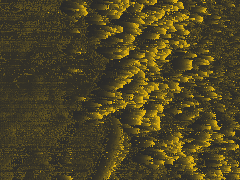} & \includegraphics[width=1in]{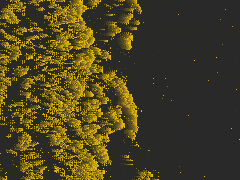} & \includegraphics[width=1in]{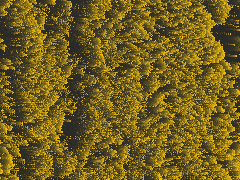} & \includegraphics[width=1in]{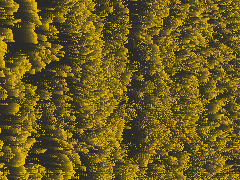}\\  
\rotatebox{90}{\textbf{Brittany}} & \includegraphics[width=1in]{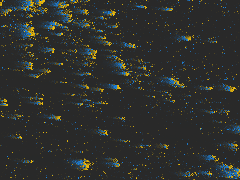} & \includegraphics[width=1in]{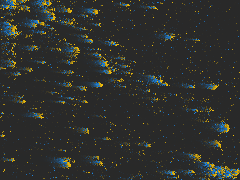} & \includegraphics[width=1in]{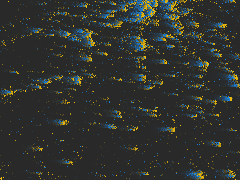} & \includegraphics[width=1in]{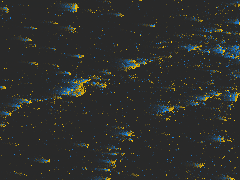} & \includegraphics[width=1in]{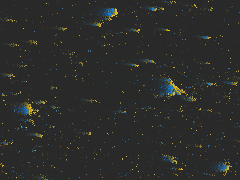}\\  
\rotatebox{90}{\textbf{Mexico}} & \includegraphics[width=1in]{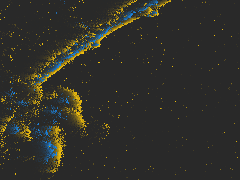} & \includegraphics[width=1in]{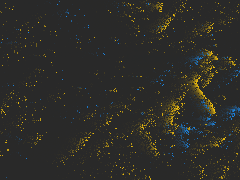} & \includegraphics[width=1in]{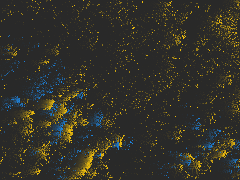} & \includegraphics[width=1in]{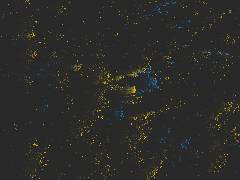} & \includegraphics[width=1in]{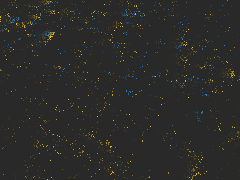}\\  
\rotatebox{90}{\textbf{Washington}} & \includegraphics[width=1in]{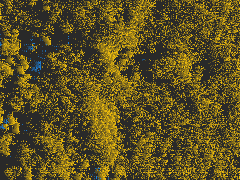} & \includegraphics[width=1in]{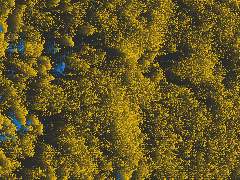} & \includegraphics[width=1in]{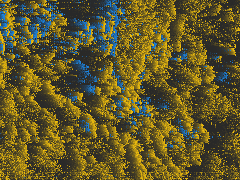} & \includegraphics[width=1in]{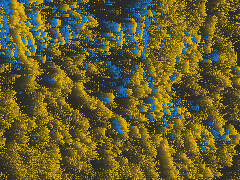} & \includegraphics[width=1in]{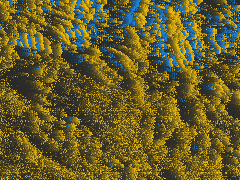}\\  
\rotatebox{90}{\textbf{Spain}} & \includegraphics[width=1in]{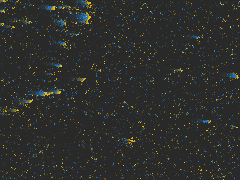} & \includegraphics[width=1in]{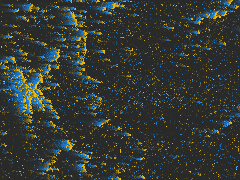} & \includegraphics[width=1in]{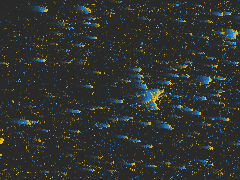} & \includegraphics[width=1in]{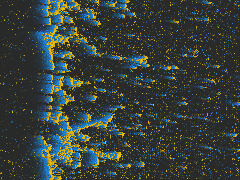} & \includegraphics[width=1in]{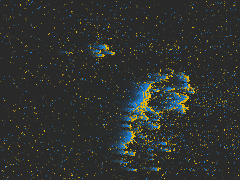}\\  
\rotatebox{90}{\textbf{Summmatra}} & \includegraphics[width=1in]{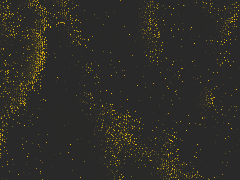} & \includegraphics[width=1in]{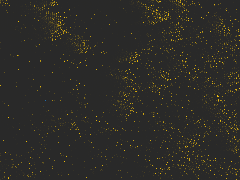} & \includegraphics[width=1in]{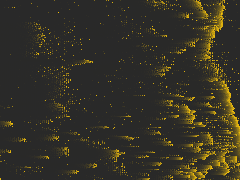} & \includegraphics[width=1in]{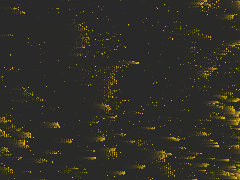} & \includegraphics[width=1in]{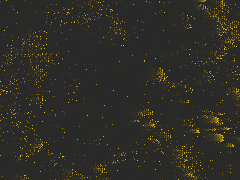}\\  
\rotatebox{90}{\textbf{UK}} & \includegraphics[width=1in]{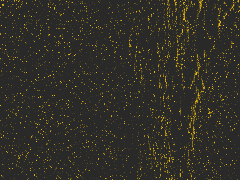} & \includegraphics[width=1in]{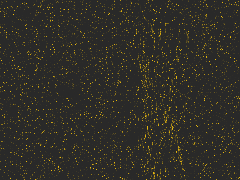} & \includegraphics[width=1in]{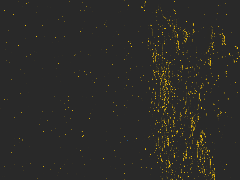} & \includegraphics[width=1in]{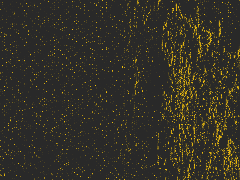} & \includegraphics[width=1in]{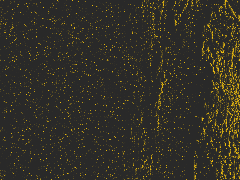}\\  
\rotatebox{90}{\textbf{Panama}} & \includegraphics[width=1in]{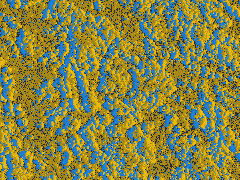} & \includegraphics[width=1in]{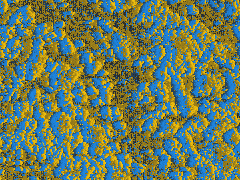} & \includegraphics[width=1in]{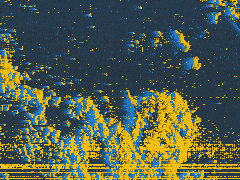} & \includegraphics[width=1in]{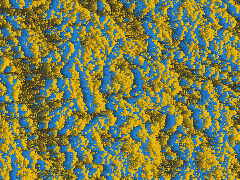} & \includegraphics[width=1in]{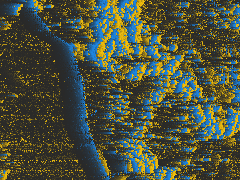}\\  
\rotatebox{90}{\textbf{Egypt}} & \includegraphics[width=1in]{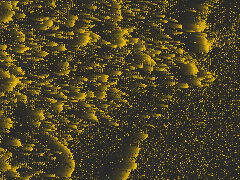} & \includegraphics[width=1in]{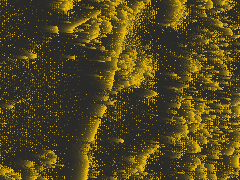} & \includegraphics[width=1in]{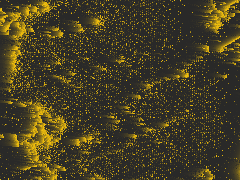} & \includegraphics[width=1in]{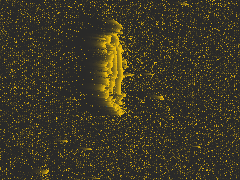} & \includegraphics[width=1in]{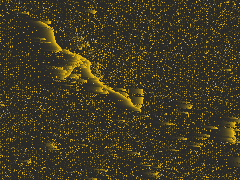}
\end{tabular} 
\caption{Selected frames from each recording in sequential order.}
\label{tab:sequencesdata}
\end{figure*}

\section{Further explanations on the One-dimensional case}

\begin{figure}[h]
  \hspace*{-1cm}
  \centering
  \includegraphics[width=6in]{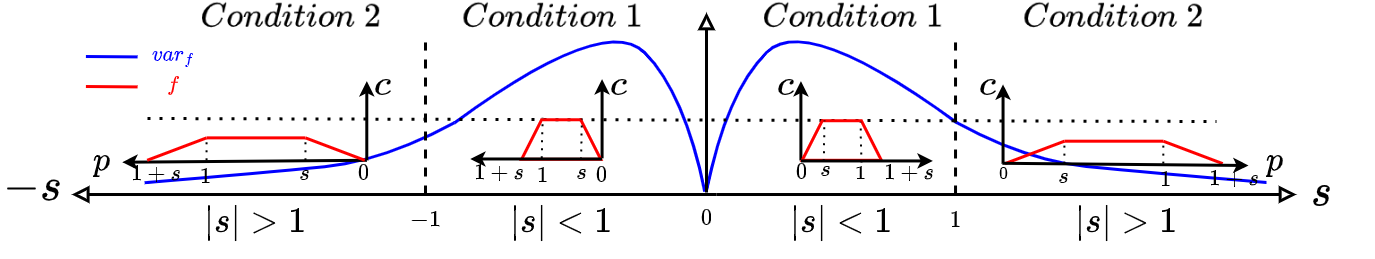}
  \caption{The changes in the geometry of the \textit{Line of Warped Events} $f$ across different $s$.}\label{fig:1dproblem}
\end{figure}

In Section 2.1, we introduced the one-dimensional correction and explained the mathematical details of the continuous noise variance for condition 1. In this section, we shall explain the details of condition 2 (figure~\ref{fig:1dproblem}).

Condition 2 starts when $s \geq 1$. It is also a piecewise linear function made of three segments, however, its height is influenced by the change of the candidate speed $s$ when it becomes larger than the pixel width. In this case, the height of the polyhedron reduces as $s$ increases. It is described as follows:

\begin{equation}
 f(p) = \begin{cases}
    \frac{cp}{s} & 0 \le p \le 1 \\
    \frac{c}{s} & 1 \le p \le s  \\
    \frac{s-p+1}{s^2} & s \le p \le 1 + s \\
   \end{cases} \label{eq:1df2_pw}
\end{equation}

Applying the formulas for the mean and variance given for $s \geq 1$:

\begin{equation}
    \overline{f} = \frac{c}{s+1} \quad \text{and}\quad \text{var}_f(s)= c^{2}\frac{s (2 - s)}{3 \left(s + 1\right)^{2}}\label{eq:varianceoutputshear}
\end{equation}

$\overline{f}$ for both conditions ($s \leq 1$ and $s \geq 1$) yield the same value, showing that the consistency of the geometrical shape is preserved even at a large speed candidate. In this case, we also want $var_f$ to be zero. We thus introduced $\alpha$ as a multiplicative correction function for the same non-constant segments.

\begin{equation}
    \lambda(p) = \alpha(p) \cdot f(p)
    \label{eq:lambda}
\end{equation}

\begin{equation}
   \alpha(p) = \begin{cases}
    \frac{s}{p} & 0 \le p \le 1 \\
    s & 1 \le p \le s  \\
    \frac{s^2}{-p + s + 1} & s \le p \le 1 + s \\
   \end{cases} \label{eq:1dcorrectionfunction}
\end{equation}

\begin{figure}[h]
  \hspace*{-1cm}
  \centering
  \includegraphics[width=6in]{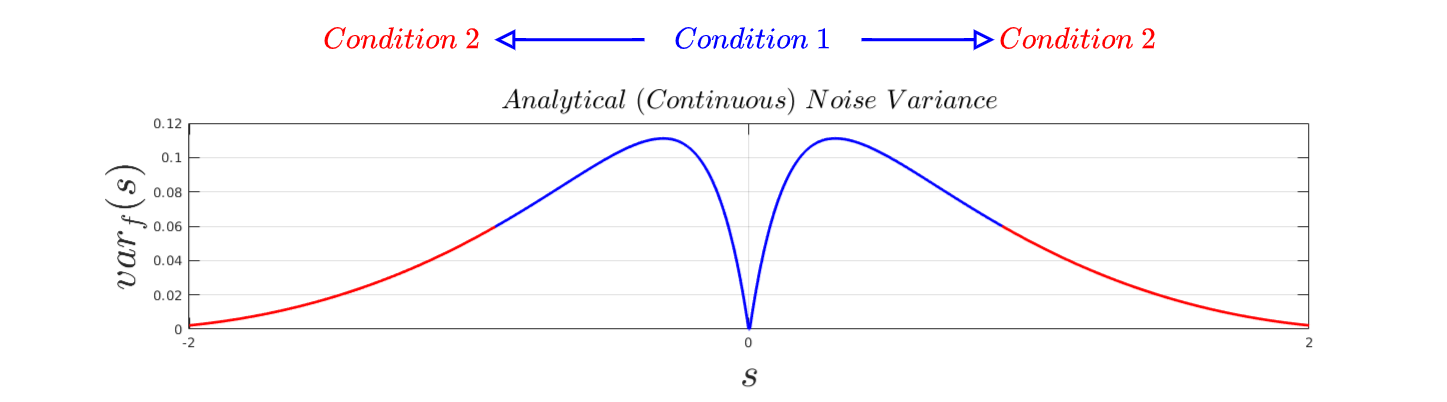}
  \caption{Plot of the variance of $f$ combining $var_f(s)$ for $s \leq 1$ and $s \geq 1$.}\label{fig:variacne2conditions}
\end{figure}

This shows that the piecewise functions for conditions 1 and 2 correctly model the changes in the geometry of the proposed solid rectangle.

\section{Further explanations on the Two-dimensional case}

In Section 2.2, we introduced the two-dimensional correction function for $s_x \geq 1$ and $s_y \geq 1$. In this section, we shall describe the entire two-dimensional space for conditions where $s_x \leq 1$ and $s_y \leq 1$.

\begin{figure}[h]
  \centering
  \includegraphics[width=5in]{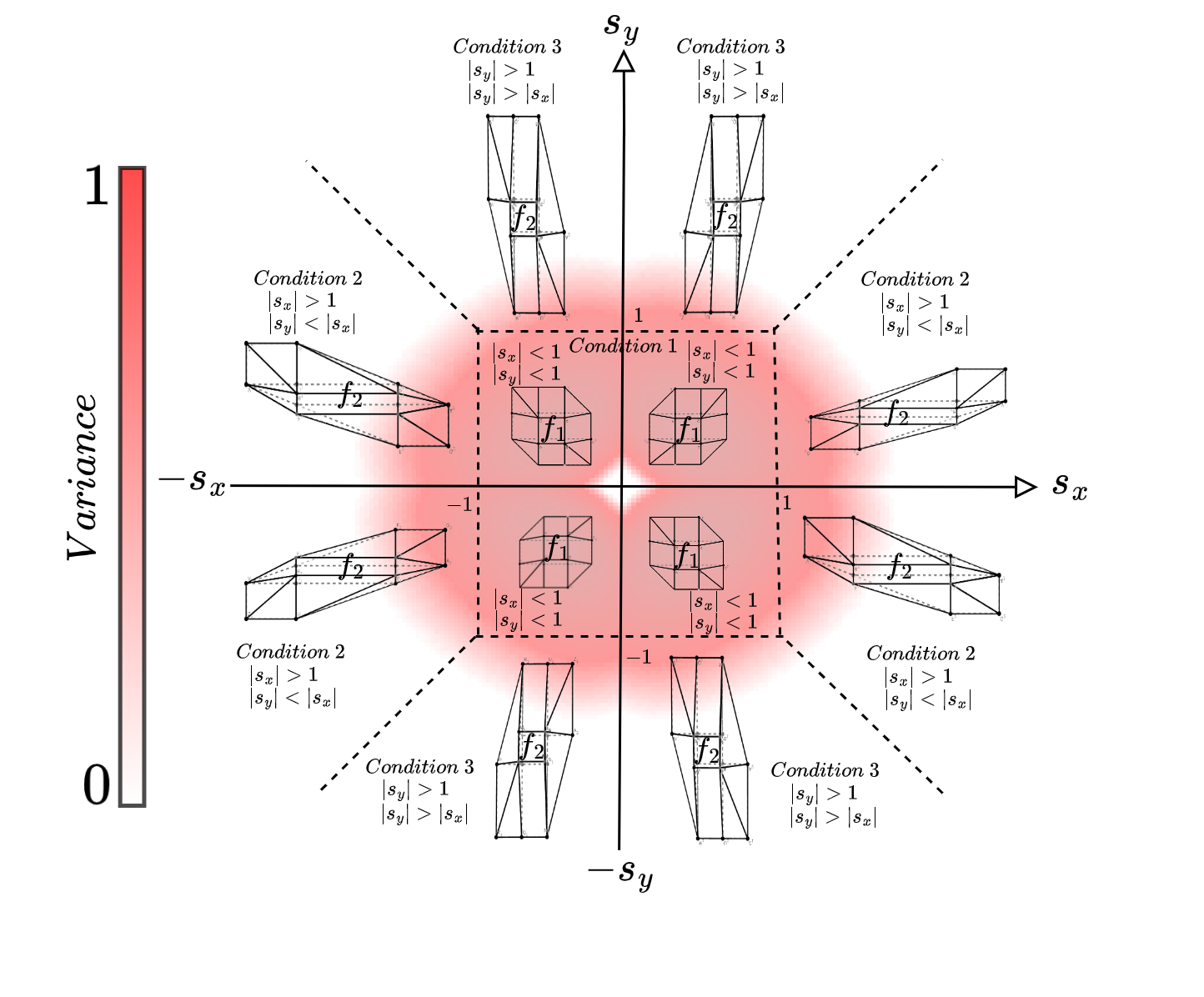}
  \caption{An illustration of the height of the sheared rectangular cuboid (accumulated warped events) as a function of $s_x$ and $s_y$ (black geometric figures) and the corresponding variance (red and white background). The problem exhibits several symmetries and can be solved by considering only two sets of conditions (condition 2 and condition 3 are symmetrical about the axis $y = x$).}
   \label{fig:geometry2d}
\end{figure}

Similarly to the one-dimensional case, once the candidate speed is greater than the width or the height of the sensor, the height of the trapezoid will start to reduce with respect to the speed parameters. As shown in figure~\ref{fig:geometry2d}, the geometry in every condition exhibits several symmetries and can be solved by considering only two sets of conditions for conditions 2 and 3.

Therefore, we define the height function $f$ for both for $s_x \geq 1$ and $s_y \geq 1$ as follows:

\begin{equation}
f(p_x, p_y) = \begin{cases}
    \frac{cp_y}{s_y} & p_x \le 1 \land p_y \le \frac{s_y}{s_xp_x} \\
    \frac{cp_x}{s_x} & p_x \le \frac{s_x}{s_yp_y} \land p_y \le \frac{s_y}{s_x}  \\
    c\left(\frac{1}{s_x}-\frac{p_x}{s_x}+\frac{p_y}{s_y} \right)  & 1 \le p_x < s_x \\ & \land \frac{(p_x - 1)s_y }{s_x} \le p_y \le \frac{p_xs_y}{s_x} \\
    \frac{cp_x}{s_x} & p_x \le 1 \land\ \frac{s_y}{s_x} \le p_y \le 1 \\
    \frac{c}{s_x} & 1 \le p_x \le s_x \\ & \land\ \frac{p_xs_y}{s_x} \le p_y \le \frac{(p_x - 1)s_y}{s_x+1} \\
    c\left(\frac{1}{s_x}-\frac{p_x}{s_x}+\frac{p_y}{s_y} \right) & s_x \le p_x \le 1+s_x \\ & \land\ \frac{(p_x - 1)s_y }{s_x} \le p_y \le s_y
   \end{cases} %
\end{equation}

Due to the consistent symmetry, calculating the mean of $f$ yields a similar value to condition 1:

\begin{equation}
    \overline{f}=\frac{c}{s_x + s_y + 1}\label{eq:2dmeanv2}
\end{equation}

\begin{equation}
    \text{var}_f(s_x, s_y)= \frac{c^{2} \left(4 s_{x}^{2} s_{y} + 4 s_{x}^{2} - 2 s_{x} s_{y}^{2} - 3 s_{x} s_{y} - 2 s_{x} + s_{y}^{2} + s_{y}\right)}{6 s_{x}^{3} \left(s_{x} + s_{y} + 1\right)^{2}}\label{eq:2dvariancev2}
\end{equation}

To cancel out the effect of noise variance, we introduce $\alpha$ that is specific to $f$ with the same properties as the previously introduced $\alpha$. That's to be multiplicative and flatten $f$ to ensure that the variance of the corrected height function is zero.

\begin{equation}
\alpha(p_x, p_y) = \begin{cases}
    \frac{s_y}{p_y} & p_x \le 1 \land p_y \le \frac{s_y}{s_xp_x} \\
    \frac{s_x}{p_x} & p_x \le \frac{s_x}{s_yp_y} \land p_y \le \frac{s_y}{s_x}  \\
    \frac{s_xs_y}{s_xp_y-s_yp_x+s_y}  & 1 \le p_x < s_x \\ & \land \frac{(p_x - 1)s_y }{s_x} \le p_y \le \frac{p_xs_y}{s_x} \\
    \frac{s_x}{p_x} & p_x \le 1 \land\ \frac{s_y}{s_x} \le p_y \le 1 \\
    s_x & 1 \le p_x \le s_x \\ & \land\ \frac{p_xs_y}{s_x} \le p_y \le \frac{(p_x - 1)s_y}{s_x+1} \\
    \frac{s_xs_y}{s_xp_y-s_yp_x+s_y} & s_x \le p_x \le 1+s_x \\ & \land\ \frac{(p_x - 1)s_y }{s_x} \le p_y \le s_y
   \end{cases} %
\end{equation}

\begin{figure}[h]
  \centering
  \includegraphics[width=4in]{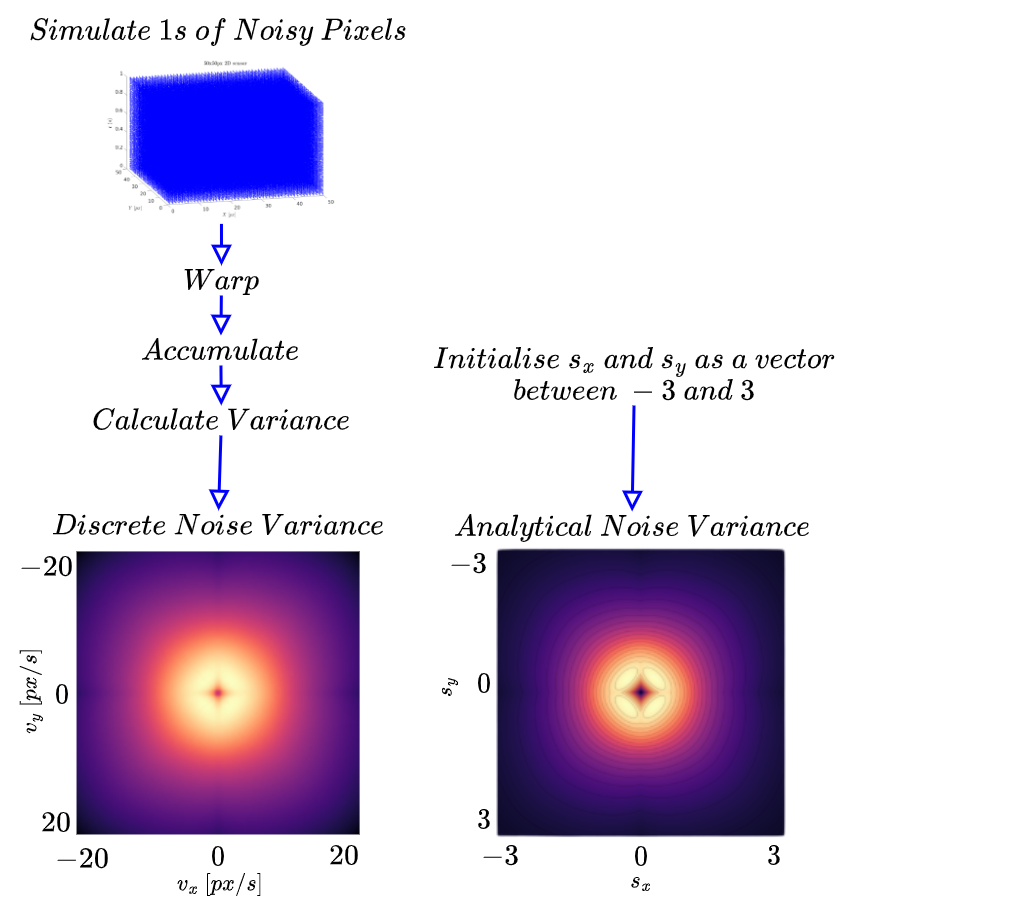}
  \caption{Comparison of the variance results between both analytical and discrete variance equation. The discrete analytical variance was performed by simulating dense noise events of dimension $50\times50$ pixels, while the analytical variance was calculated using $\text{var}_f(s_x, s_y)$.}
   \label{fig:geometry2dcomparison}
\end{figure}

In figure~\ref{fig:geometry2dcomparison}, we compared the output variance using the analytical formula as a function of the candidate speed and the discrete simulated noise. Both show an exact match.

\addtolength{\textheight}{-12cm}